\documentclass[runningheads]{llncs}

\usepackage{eccv}

\usepackage{eccvabbrv}

\usepackage{graphicx}
\usepackage{booktabs}
\usepackage{multirow}
\usepackage{colortbl}
\usepackage{listings}
\usepackage{adjustbox}
\usepackage{xcolor}
\usepackage{enumitem}
\usepackage{pifont}
\usepackage{cuted}
\usepackage{svg}
\usepackage{amsmath}
\usepackage{wasysym}
\usepackage[misc]{ifsym}
\usepackage{wrapfig}
\usepackage{pifont}
\usepackage[accsupp]{axessibility}  

\usepackage{hyperref}

\usepackage{orcidlink}

\begin{document}

\title{ReViP: Mitigating False Completion in Vision-Language-Action Models with Vision-Proprioception Rebalance} 

\titlerunning{ReViP: Mitigating False Completion in VLA Models}


\author{Zhuohao Li\inst{1,2}\orcidlink{0009-0000-9436-8794}
\and
Yinghao Li\inst{1,2}\orcidlink{0009-0008-5835-6655}
\and
Jian-Jian Jiang\inst{1}\orcidlink{0009-0009-2324-639X}
\and
Lang Zhou\inst{1,2}\orcidlink{0009-0006-6911-5078}
\and
Tianyu Zhang\inst{3,2}
\and
Jiadong Yin\inst{1}\orcidlink{0009-0005-2785-2945}
\and
Mu Lin\inst{1}\orcidlink{0009-0005-6021-5630}
\and
Yi-Lin Wei\inst{1}\orcidlink{0009-0004-9210-7370}
\and
Wei-Shi Zheng\inst{1,4,5,6 *\orcidlink{0000-0001-8327-0003}}}

\authorrunning{Z. Li~et al.}

\institute{Sun Yat-sen University, China \and
Shenzhen Loop Area Institute, China \and
Beijing Institute of Technology, China \and 
Peng Cheng Laboratory, Shenzhen, China \and 
Key Laboratory of Machine Intelligence and Advanced Computing, Ministry of Education, China \and 
Guangdong Province Key Laboratory of Information Security Technology, China\\
\email{lizhh268@mail2.sysu.edu.cn; wszheng@ieee.org}
}


\begingroup
\renewcommand{\thefootnote}{\relax}
\footnotetext{*: corresponding author.}
\endgroup
\maketitle

\begin{center}
Project Page: \url{https://isee-laboratory.github.io/ReViP/}
\end{center}

\begingroup
\centering
\includegraphics[width=0.95\textwidth]{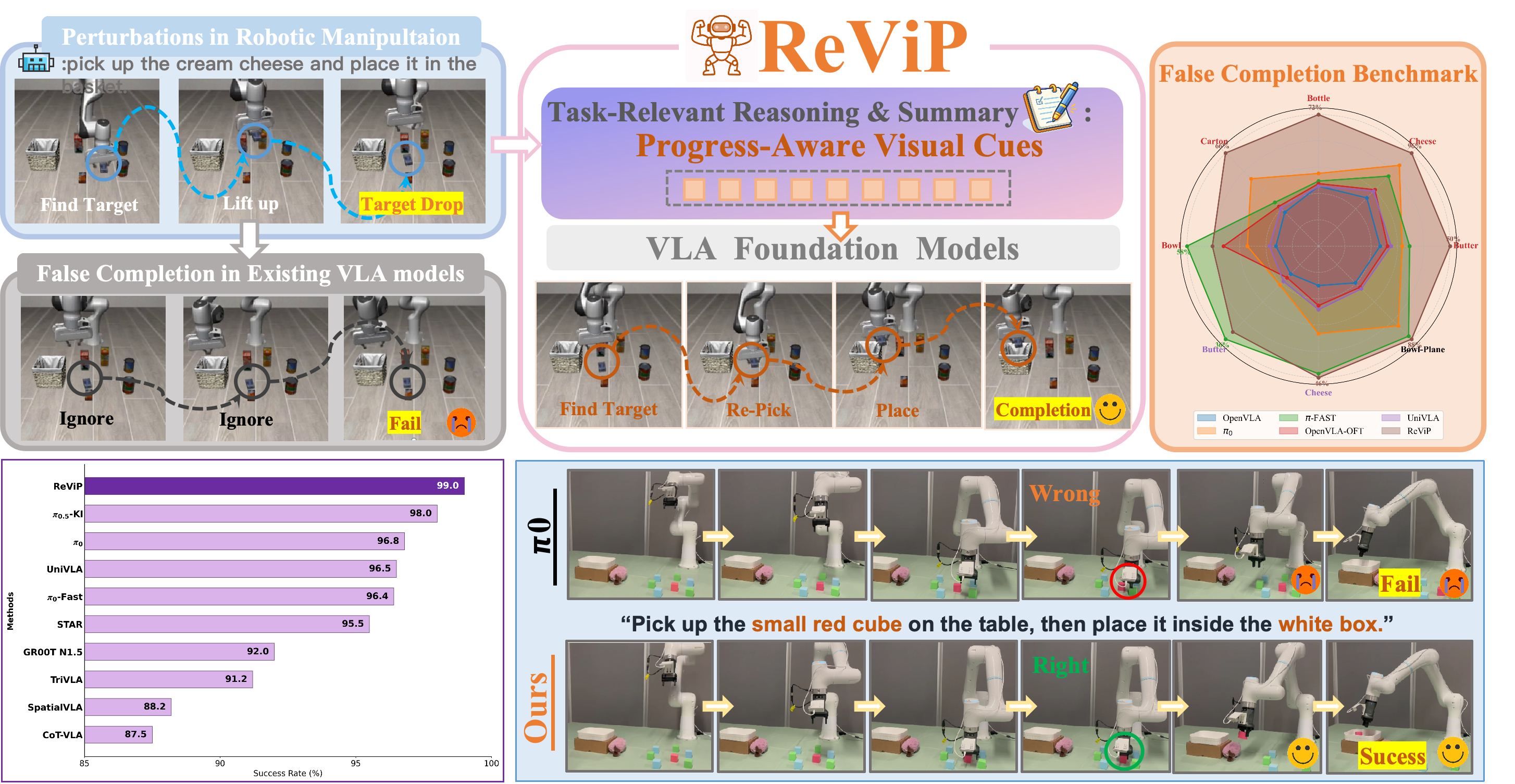}\par
\captionsetup{type=figure}
\caption{\textbf{From False Completion to Vision-Proprioception Rebalance.} 
\textbf{Top:} \emph{False completion} in VLA models occurs when policies prematurely fail despite unmet goals, often due to over-prioritizing internal state progression over visual feedback.
\emph{ReViP} addresses this by injecting progress-aware visual cues, rebalancing visual and proprioceptive dynamics for successful task completion;
To evaluate and reproduce false completion, we introduce the \textbf{first False-Completion Benchmark Suite}, featuring eight tasks with controlled perturbations such as \emph{Object Drop}, \emph{Distractor Swap}, and \emph{Relayout}.
\textbf{Bottom:} 
Extensive experiments in both \emph{simulation} (False-Completion, LIBERO, RoboTwin 2.0) and \emph{real-world} settings show that ReViP significantly reduces false completion and improves task success rates, consistently outperforming existing methods.
}
\label{fig:setting}
\endgroup


\begin{abstract}
Vision-Language-Action (VLA) models have advanced robotic manipulation by combining vision, language, and proprioception to predict actions.
However, previous methods fuse proprioceptive signals directly with vision-language features, resulting in state-dominant bias and \textbf{false completions} despite visible execution failures.
We systematically analyze this failure mode, attributing it to modality imbalance, where policies overly rely on internal state progression and underuse visual evidence.
To address this, we introduce the first \textbf{False-Completion Benchmark Suite}, featuring eight tasks with three controlled perturbations (\emph{Object Drop}, \emph{Distractor Swap}, \emph{Relayout}) to comprehensively evaluate false completion.
Moreover, we propose \textbf{ReViP}, a novel VLA framework with \textbf{Vi}sion-\textbf{P}roprioception \textbf{Re}balance to enhance visual grounding and robustness under perturbations. 
The key insight is to introduce auxiliary \emph{progress-aware visual cues} to adaptively modulate the coupling between semantic perception and proprioceptive dynamics.
Specifically, progress-aware visual cues are extracted by an external Task-Stage Observer, which performs task-relevant reasoning on real-time observations to drive task-stage feature-wise linear modulation, enhancing environmental awareness and mitigating state-driven errors.
Extensive experiments show that ReViP effectively mitigates false completion and improves success rates over strong VLA baselines, achieving a \textbf{26\%} gain over $\pi_0$ model on our suite, with gains extending to LIBERO, RoboTwin 2.0, and real-world evaluations.

\end{abstract}    
\section{Introduction}
\label{sec:intro}
Recently, Vision-Language-Action (VLA) models~\cite{Rt-2,kim2024openvla,spatialvla,fast,hou2025dita, pi05, huang2025thinkact,zhong2025flowvla} have demonstrated impressive progress in robotic manipulation. 
Despite encouraging progress, previous methods typically encode visual-language inputs via Visual-Language Models (VLMs)~\cite{Qwen2.5-VL,liu2023llava,chen2024internvl,zhong2025dexgraspvla,li2024generalist_vla,zhai2025vla_critic} and directly fuse proprioceptive signals~\cite{black2024pi_0, oft}, leading to state-dominant bias and false completions even in the presence of clear visual failure cues. 
We attribute this behavior to modality imbalance, where policies over-rely on internal state while underusing visual evidence.
For instance, when a target object drops during the execution, the policy continues the planned placement action towards the goal region, despite the target remaining visible and requiring retrieval. 
As shown in Fig.~\ref{fig:setting}, the policy may prematurely halt or incorrectly declare success when the goal is unmet, causing the robot to terminate the operation without further action. We term this phenomenon \textbf{``False Completion''}.
This behavior suggests that visual feedback is underutilized, with decision-making processes relying on proprioceptive cues, which conflicts with human common-sense task completion reasoning.

To investigate this, we first conduct systematic modality-controlled studies to comprehensively understand false completion and its underlying causes.
We find that naively removing state inputs reduces performance, highlighting the necessity of rebalancing visual and proprioceptive signals.
To address the lack of a benchmark for false completion, we then introduce the first \textbf{False-Completion Benchmark Suite} on the LIBERO~\cite{libero}, enabling evaluation across VLA models.
The suite features eight distinct tasks and three controlled perturbation sources (\emph{Object Drop}, \emph{Distractor Swap}, and \emph{Relayout}), enabling a comprehensive evaluation of model performance in response to false completion across a diverse set of perturbations.
Each perturbation addresses a unique aspect of model robustness: \emph{Object Drop} evaluates the model's response to dynamic disruptions, while \emph{Distractor Swap} challenges its capacity to distinguish between targets and distractors. 
And \emph{Relayout} tests the model's adaptability to changes in the spatial configuration of the environment.

Additionally, to address false completion arising from modality imbalance, we present ReViP, a VLA framework that enhances visual grounding and robustness under perturbations by rebalancing vision perception with proprioceptive dynamics.
To be specific, ReViP constructs a \textbf{Task-Stage Observer} based on an external VLM to process the current observation and instruction and perform task-relevant reasoning to extract \textbf{progress-aware visual cues} that reflect task progress and environmental state. 
Building on this, the \textbf{Task-Stage Enhancer} is proposed to inject these priors into the policy via a \emph{Vision-Proprioception Feature-wise Linear Modulation}, adaptively rebalancing semantic perception and proprioceptive dynamics at the feature level.
This design not only promotes feedback awareness but also mitigates state-driven errors under perturbations, such as object drops, distractor-induced mismatches, and environment relayout. 

Finally, we conduct extensive evaluations across multiple simulation benchmarks, including false-completion assessment, the standard LIBERO, and the dual-arm RoboTwin 2.0 benchmark \cite{chen2025robotwin}, spanning a diverse set of robotic manipulation tasks. 
In addition, real-world experiments are performed to further validate our hypothesis. The results reveal that ReViP significantly reduces false completions under previously unseen disturbances and achieves state-of-the-art (SOTA) task success rates, consistently outperforming existing methods on these challenging tasks.
We will open-source our code and benchmark for the development of the community. Our contributions can be summarized as follows:
\begin{itemize}
\item We identify \emph{false completion} as a critical failure mode in VLA models, arising from modality imbalance which leads to state-dominant bias.
We systematically study this phenomenon through modality-controlled experiments to reveal its underlying causes.

\item We introduce the \emph{first False-Completion Benchmark Suite} tailored for comprehensively evaluating model performance under diverse perturbations (\emph{Object Drop}, \emph{Distractor Swap}, \emph{Relayout}), emphasizing robustness across various false-completion settings.

\item We propose \emph{ReViP}, inspired by vision-proprioception rebalance, integrating a Task-Stage Observer for progress-aware feedback and a Task Stage Enhancer to adaptively strengthen the visual stream at the feature level.

\item We demonstrate that ReViP significantly mitigates false completions and achieves SOTA task success rates, consistently outperforming strong VLA baselines across LIBERO, RoboTwin 2.0, and real-world tasks.
\end{itemize}

\section{Related Work}
\label{sec:related_work}

\noindent\textbf{Vision-Language-Action Models.}  VLA models~\cite{Rt-2,black2024pi_0,fast,hou2025dita,yang2025instructvla, zhou2025opendrivevla, li2025cronusvla, zhang2025dreamvla} connect perception to control by mapping pretrained vision-language representations to executable actions. 
Early approaches in the RT line (RT-1~\cite{Rt-1}, RT-2~\cite{Rt-2}, RT-H~\cite{Rt-h}) introduced action tokenization and trained transformer policies from scratch on large-scale demonstration corpora such as Open X-Embodiment~\cite{OXE} and DROID~\cite{khazatsky2024droid}. 
Recently, the $\pi$ series (FAST \cite{fast}, $\pi_0$ \cite{black2024pi_0}, and $\pi_{0.5}$ \cite{pi05}) proposed heterogeneous co-training across robots and used flow matching-based decoders \cite{flowmatching} for continuous action generation. 
Beyond these, a range of VLA variants~\cite{cotvla,spatialvla,star,trivla,hou2025dita,li2025cogvla,yang2026shakevla} explore performing better robot manipulation, including structured reasoning, spatial grounding, and efficient architectures.
For example, CoT-VLA~\cite{cotvla} integrates chain-of-thought style reasoning with a lightweight policy to improve efficiency and the GR00T series \cite{gr00t} targets generalist humanoid robots by developing open foundation VLA models for complex bimanual and whole‑body tasks.
Few studies explicitly explore the modality mechanism in VLA models \cite{zhao2025nostate}, and the phenomenon of false completion remains underexplored.
In contrast, in this work, we focus on this modality imbalance and target false completion by explicitly rebalancing visual and proprioceptive signals.

\noindent\textbf{VLA Models with External VLMs.}
Directly fine-tuning the VLM inside a VLA on robot data often degrades its pretrained multimodal understanding because of distribution shift and forgetting \cite{black2024pi_0, pi05, chatvla2}, which motivates keeping a strong external VLM for task understanding \cite{actionslanguagefinetuningvlms, li2024cogact}. 
Recent methods fall into two patterns. The first uses the VLM to decompose long-horizon instructions into subgoals, leaving the VLM only as a planner without exposing its internal semantic features to the VLA backbone \cite{ahn2022icanisay, innermonologueembodiedreasoning}. 
The second uses the VLM as a goal or step judge to decide success and trigger handcrafted retries or resets, which requires task-specific flows and incurs expensive restarts \cite{vlsuccessdetectors,duan2024aha}.
While generic external success verification remains underexplored, frameworks like VLATest \cite{wang2025vlatest} use auxiliary evaluators to probe and stress-test VLA behavior across variations.
Differing from them, ReViP injects internal VLM representations into the VLA via task-stage modeling feedback to rebalance vision and proprioception.

\section{False Completion}
\label{sec:FC}
In this section, we first define false completion, a failure mode where VLA policies prematurely proceed or terminate despite unmet task goals. We then explore the impact of modality imbalance through two key questions, and introduce our False-Completion Benchmark.

\subsection{Problem Setup}
\label{sec:definition}
\noindent\textbf{False completion definition.}
Here, we formalize the setting of false completion. The observation $I_t$ at time step $t$ consists of synchronized images from a set of cameras $\mathcal{C}$ that include egocentric and exocentric views:
\begin{equation}
    I_t \;=\; \bigl\{\,I_t^{(c)}\in\mathbb{R}^{H_c\times W_c\times 3}\ \big|\ c\in\mathcal{C}\,\bigr\}.
\end{equation}
Given $(I_t, S_t, l)$, where $S_t$ is proprioception and $l$ is the language instruction, a policy $\pi_{\theta}$ then predicts a block of $n$ future actions (action chunk) as follows:
\begin{equation}
\begin{alignedat}{2}
        A_t = \pi_{\theta}\bigl[a_t, a_{t+1}, \dots, a_{t+n-1} \mid (I_t, S_t, l)\bigr]
\end{alignedat}
\end{equation}
Then, a termination flag $d_t\in\{0,1\}$ is defined, where $d_t=1$ is a completion declaration by the policy.
Note that $d_t=1$ is a conceptual indicator, reflecting a lack of significant action change over time, rather than a direct decision from the model itself. 
Let $G(I_t)\in\{0,1\}$ be a visual goal predicate decided from the multi-view images. Following that, a task instance is considered truly completed only when both the policy declares completion and the visual predicate confirms goal satisfaction.
However, when the policy issues a completion declaration while a visual predicate indicates the goal is unmet, we obtain a \textbf{false completion}:
\begin{equation}
\begin{alignedat}{2}
\text{True Completion} \;&=\; \!\left[d_t=1\ \wedge\ G(I_t)=1\right], \\
\text{False Completion} \;&=\; \!\left[d_t=1\ \wedge\ G(I_t)=0\right].
\end{alignedat}
\end{equation}

\subsection{False Completion in VLAs}
In this paper, we attribute this failure mode to a modality imbalance within VLA models. Specifically, we identify a \textbf{state-dominant bias}: a tendency for the policy to prioritize internal proprioceptive sequences over external perceptual feedback. This bias causes the robot to blindly persist in task execution even when visual evidence signaling failure (e.g., an object being dropped) is present, ultimately leading to false completions.

Building on our hypothesis, to understand the effect of modality imbalance, we aim to answer two questions: 
\textbf{Q1:} \textit{Is false completion caused by modality imbalance (state-dominant bias)} and \textbf{Q2:} \textit{Can false completion be mitigated by simply removing state inputs}.
To answer these, we conduct targeted modality-controlled experiments on both real robots.

\begin{wraptable}{r}{0.54\columnwidth}
\centering
\vspace{-8mm}
\caption{Real-world modality-controlled behavior. \#Proceed-to-Goal and \#Back-to-Object reflect the policy's response to a state-vision mismatch with Success Rate for normal deployment.}
\label{tab:modality}
\scriptsize
\setlength{\tabcolsep}{3pt}
\resizebox{\linewidth}{!}{%
\begin{tabular}{cccc}
    \toprule
    \textbf{Modality} & \textbf{\#Proceed-to-Goal $\downarrow$} & \textbf{\#Back-to-Object $\uparrow$} & \textbf{Success Rate (Normal) $\uparrow$} \\
    \midrule
    State-Enabled & 46 / 50 & \phantom{0}4 / 50 & \textbf{70\%} \\
    State-Masked & \textbf{17 / 50} & \textbf{33 / 50} & 40\%\\
    \bottomrule
\end{tabular}%
}
\vspace{-5mm}
\end{wraptable}

\noindent\textbf{(1) Modality imbalance leads to false completion.}
We first investigate whether false completion arises from state-dominant modality imbalance by reproducing a state–vision mismatch in a real-world \emph{cup pick-and-place} task (detailed in Section~\ref{sec: real-world}).
By prematurely closing the gripper near the object, we induce a state-vision mismatch: the state signal indicates the object ``grasped'' while the object is not visually grasped.
In over 50 experimental trials, we record the robot's behavioral response, categorizing actions as either \emph{Proceed-to-Goal} or \emph{Back-to-Object}.
We mask proprioceptive signals by zeroing out the state input (\emph{masking state}), forcing the model to rely solely on visual inputs.

As shown in Table~\ref{tab:modality}, with state enabled (i.e., $\pi_0$), the robot predominantly follows proprioception, proceeding to the goal (\emph{46 out of 50}) despite the visual mismatch, exhibiting false completion.
This demonstrates that false completion is primarily driven by over-reliance on proprioceptive signals, i.e., state-dominant bias.

\noindent\textbf{(2) Naively removing state helps false completion at a cost.}
Building on the above analysis that state-dominant bias causes false completion, we further explore whether removing state inputs can mitigate the issues.
Table~\ref{tab:modality} shows that when state signals are masked, the robot exhibits markedly different behavior, returning to the object in 33 out of 50 trials. This confirms that removing state inputs increases reliance on vision and partially alleviates false completion.
However, naive removal comes at a cost: overall success during standard task execution drops sharply from 70\% to 40\%, demonstrating that proprioceptive signals provide critical task-relevant information that vision alone cannot recover.
\textbf{Naively removing state inputs fundamentally compromises the model's capabilities, highlighting the need to rebalance visual and proprioceptive signals},
which motivates the design of ReViP (Section~\ref{sec:Method}).

We also conduct modality-ablation experiments on simulation benchmarks to further validate our findings. Results are consistent with the real-world study, and detailed settings are provided in the supplementary materials.

\subsection{False-Completion Benchmark Suite}
\label{sec: FC benchmark}

\begin{wrapfigure}[22]{r}{0.5\linewidth}
    \vspace{-8mm}
    \centering
    \includegraphics[width=\linewidth]{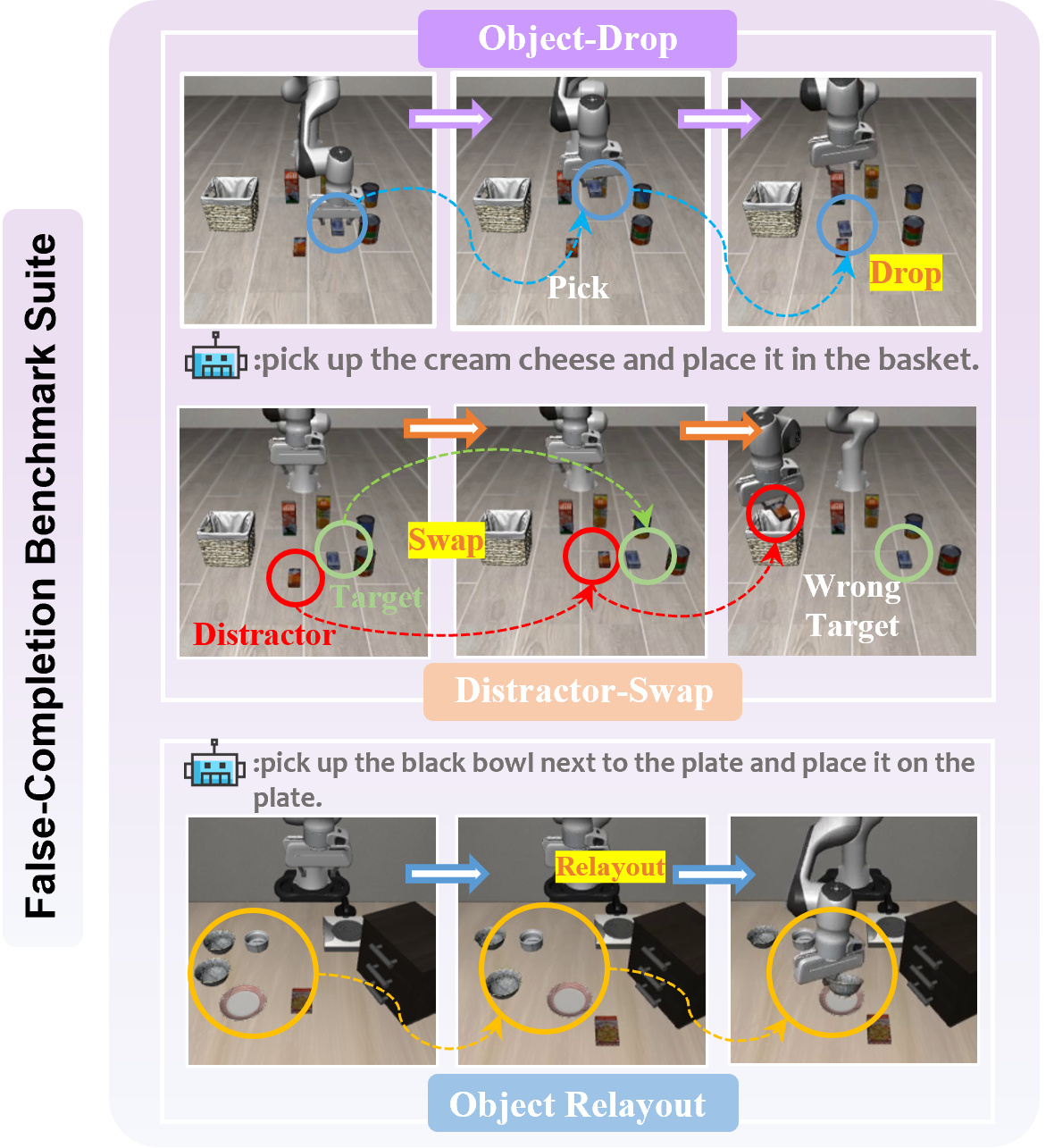}
    \caption{Illustrative examples of the False-Completion Benchmark. Object Drop tests response to execution-time failures. Distractor Swap tests instance-level grounding under similar objects. Relayout tests spatial adaptation when targets and goals appear in new configurations.}
    \label{fig:FC}
    \vspace{-6mm}
\end{wrapfigure}

To comprehensively evaluate false completion and in the absence of a suitable benchmark, we introduce the \textbf{False-Completion Benchmark Suite} on LIBERO.

Its design principle is simple: keep the instruction fixed while perturbing the scene at initialization or during execution, so that replaying demonstration-like proprioceptive trajectories is insufficient and policies must rely on current visual evidence.

The suite contains eight distinct tasks from three perturbation families (Fig.~\ref{fig:FC}), each probing a different failure surface of state-dominant policies.
All perturbations are deterministic (including a reproducible drop trigger) to ensure fair model comparisons. Full details of the benchmark settings are provided in the supplementary materials.

\noindent\textbf{Object Drop.}
This family directly stresses recovery from unexpected displacement (objects of varying sizes and cabinet-contact cases), where policies must detect failure and regrasp instead of continuing the original placement trajectory.

\noindent\textbf{Distractor Swap.}
This family swaps target and distractor poses while keeping language unchanged, forcing instance-level visual grounding and exposing failures caused by state-driven trajectory replay.

\noindent\textbf{Object Relayout.}
This family jointly relocates the target and goal region, breaking demonstration-specific spatial priors and requiring replanning from current visual configuration.

In summary, the benchmark treats false completion as a primary robustness challenge, with controlled perturbations that diagnose visual grounding, progress monitoring, and recovery behavior in a unified setting.



\section{ReViP}
\label{sec:Method}
To address false completion arising from state-dominant bias, we design ReViP, whose core idea is to rebalance visual and proprioceptive signals.
Simply removing state inputs does not resolve this bias and can degrade task performance, motivating a more careful feature-level modulation.
In this section, we introduce the Task-Stage Observer (TSO), which extracts progress-aware visual cues (Section \ref{sec:observer}), followed by the Task-Stage Enhancer (TSE), which adaptively rebalances visual-proprioceptive signals (Section \ref{sec:enhancer}). 
We then describe the action prediction process (Section \ref{sec:action}) and summarize the overall structure of ReViP.

\begin{figure*}[ht]
    \centering
    \includegraphics[width=0.92\linewidth]{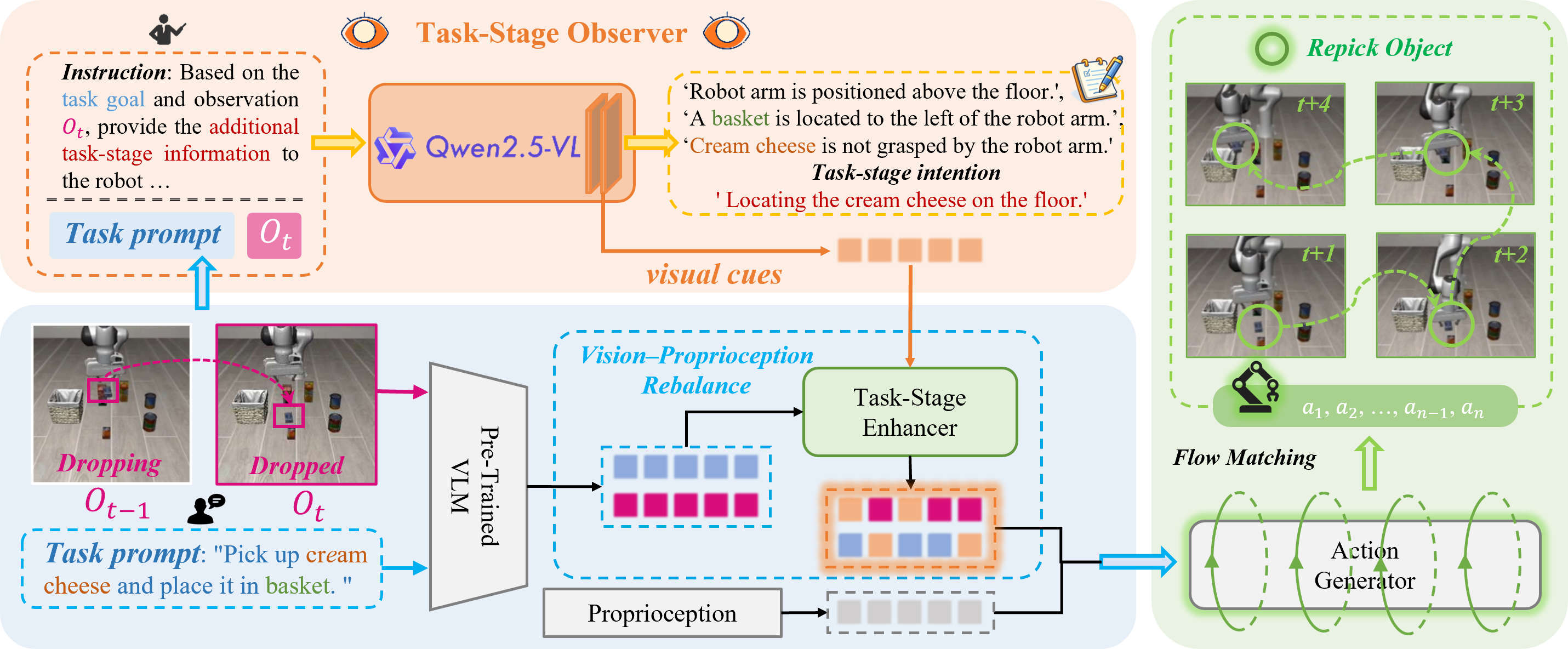}
    \caption{Overview of ReViP. It consists of two stages: (above) Task-Stage Observer for cues extraction and (below) Task-Stage Enhancer for rebalancing. Through a frozen vision-language model, progress-aware visual cues are extracted from the observation $O_t$ and task instructions. These cues are then injected into the VLA backbone by the Task-Stage Enhancer to adaptively rebalance the visual and proprioceptive streams.}
    \label{fig:ReViP}
    \vspace{-12mm}
\end{figure*}

\subsection{Task-Stage Observer}
\label{sec:observer}
To deal with visual information being underused, we propose TSO to convert observation $I_t$ and instruction $l$ into task-centric visual cues while explicitly filtering task-irrelevant content.
We employ a Qwen 2.5-VL~\cite{Qwen2.5-VL} as the representative vision-language backbone for the TSO.

At time step $t$, TSO takes ($I_t$, $l$) as inputs, and performs task-relevant reasoning that first identifies the robot's currently visible physical state and the spatial location and status of the task-relevant objects, then summarizes the immediate stage intention aligned with the given goal.
The resulting cues provide explicit visual evidence for decision-making and generate a concise hypothesis for the subsequent task stage.
As illustrated in the representative \emph{object drop} scenario (Fig.~\ref{fig:ReViP}), when the target \emph{cream cheese} accidentally falls while the robot is moving toward the basket from $I_{t-1}$ to $I_t$, TSO analyzes $(I_t, l)$ to summarize critical task-stage evidence such as \emph{“Cream cheese is not grasped by the robot arm”}. 
These cues indicate the target is no longer held by the gripper and localize it within the scene.
By localizing the dropped object and generating a new stage intention, e.g., \emph{“Locating the cream cheese on the floor.”}, the TSO guides the policy to initiate a re-pick action instead of continuing toward the basket.
%

To inject these discrete language cues into the VLA backbone, we adopt an LLM-based embedding extraction strategy~\cite{tang2025understanding, tao2025llmseffectiveembeddingmodels, jiang2024vlm2vec, meng2025vlm2vecv2, li2025functional} that converts the last layer hidden states into a compact continuous feature:
\begin{equation}
z_t \;=\; \mathrm{Proj}\;\!\Big(\mathrm{Pool}\big(\mathrm{TSO}_{L}(I_t,\,l)\big)\Big)\in\mathbb{R}^{d},
\end{equation}
where $L$ denotes the last layer of the external VLM, $\mathrm{Pool}(\cdot)$ aggregates the selected token representations into a single vector, and $\mathrm{Proj}(\cdot)$ is a linear projection that maps this vector to the VLA semantic space of dimension $d$.

\subsection{Task-Stage Enhancer}
\label{sec:enhancer}
Building upon the TSO, we introduce TSE to strengthen task-relevant visual features and to rebalance the relative influence of vision and proprioception during action generation.

The TSE first translates the extracted task-relevant cues $z_t$ into adaptively controlled signals via TS-FiLM. This mechanism ensures that decision-making is driven by current task-centric visual cues rather than an inertial reliance on internal state, which directly targets the false-to-true completion shift.
Specifically,  given $z_t$, the TSE produces feature-wise modulation parameters through a compact bottleneck mapping $h(\cdot)$, which can be expressed as:
\begin{equation}
[\gamma_t,\ \beta_t] \;=\; h(z_t),
\end{equation}
where $\gamma_t$, $\beta_t$$\in\mathbb{R}^{D}$, and $D$ is the token hidden size of the VLM backbone in VLA models.
In the ReViP framework, let \(P_t\in\mathbb{R}^{B\times S\times D}\) represent the concatenated vision and language prefix tokens, accompanied by a binary validity mask \(M_t\in\{0,1\}^{B\times S}\).
We then apply token-wise TS-FiLM to the vision-language prefix before action generation.
Therefore, TSE generates modulated prefix tokens $\tilde P_t$ before action prediction:
\begin{equation}
\tilde P_t \;=\; \Big(P_t + \alpha\big(\gamma_t \odot P_t + \beta_t\big)\Big)\odot M_t,
\end{equation}
where \(\alpha\in\mathbb{R}_{+}\) is a learnable modulation factor and \(\odot\) denotes Hadamard product over tokens.
This operation performs feature-level injection of progress-aware visual cues, effectively amplifying channels aligned with visual evidence while attenuating distractors that contribute to state-dominant bias.

\subsection{Action Prediction.}
\label{sec:action}
The action head is trained to predict a conditional velocity field $v_\theta\big(.\ ,\ . \big|\ \tilde F_t\big)$ via flow matching on the modulated prefix $\tilde F_t$. For a ground-truth action chunk \(A_t\) and random noise \(\varepsilon\), we sample \(\tau\sim\mathcal{U}(0,1)\) and construct a noisy interpolation along the straight-line path:
\begin{equation}
    v_\tau \;=\; (1-\tau)A_t+\tau\,\varepsilon,
\end{equation}
The target velocity (i.e., the time derivative of this path) is:
\begin{equation}
    u_{\tau} \;=\; \frac{d v_{\tau}}{d \tau} \;=\; \varepsilon - A_t,
\end{equation}
Above all, the training objective regresses the predicted velocity onto this target:
\begin{equation}
\begin{alignedat}{2}
\tilde F_t \;&=\; Fusion \; \big(\tilde P_t, S_t\big), \\ 
\mathcal{L}_{\mathrm{FM}} \;&=\; \big\|\,v_\theta\big(v_\tau,\ \tau\ \big|\ \tilde F_t\big)\ -\ u_\tau\big\|_2^2.
\end{alignedat}
\end{equation}
When a perturbation such as an object drop occurs, $z_t$ steers [$\gamma_t$, $\beta_t$] to emphasize evidence and suppress irrelevant content, which guides the action generator to approach and regrasp the object instead of proceeding to the goal, reducing false completion.

\noindent \textbf{Overall Framework.}
Integrating these components, ReViP processes the observation and instruction with the TSO, distilling them into progress-aware visual cues that capture relevant objects and stage intent. TSE injects these cues into the VLA backbone via TS-FiLM, enhancing VLM-derived features while adaptively rebalancing visual and proprioceptive signals. Finally, the modulated representation conditions a flow matching-based action generator to predict action chunks in parallel.

\section{Experiments}

\label{sec:Experiments}

\subsection{Simulation Benchmark Experiment.}
\label{sec:simulation benchmark}
We evaluate ReViP in two key aspects: First,  we test robustness using our False-Completion Benchmark Suite, which applies controlled perturbations to induce state-dominant bias. Second, we assess generalization on the LIBERO \cite{libero} and RoboTwin 2.0 benchmark \cite{chen2025robotwin}, covering diverse scenes and tasks.

\noindent\textbf{Implementation Details.} 
Training is performed on 8 $\times$ H100 GPUs (80GB) with a total batch size of 32 for 60k training steps.
ReViP uses $\pi_0$, a representative model, as the backbone policy, with TSO instantiated using Qwen2.5-VL-3B (\textbf{ReViP}).
We also evaluate a stronger observer with Qwen2.5-VL-72B (\textbf{ReViP*}) to study how VLM capacity affects cue quality, providing a comprehensive validation mechanism.
Note that TSO runs asynchronously, triggered once per action chunk, without disrupting real-time closed-loop control.
Episodes succeed only if the instructed goal is reached within the step budget.
Perturbation triggers are deterministic (e.g., reproducible drop rule) for fair comparison. Full evaluation protocols are provided in the supplementary materials.



\noindent\textbf{Baselines.} We compare ReViP with representative VLAs, including OpenVLA \cite{kim2024openvla}, OpenVLA-OFT \cite{oft}, SpatialVLA \cite{spatialvla}, $\pi_0$ \cite{black2024pi_0}, $\pi_0$-Fast \cite{fast}, CoT-VLA \cite{cotvla}, TriVLA \cite{trivla}, GR00T N1.5 \cite{gr00t}, and UniVLA \cite{univla} in single-arm settings. 
For bimanual-arm setting, we further evaluate DP3 \cite{Ze2024DP3}, RDT \cite{liu2025rdtb} and $\pi_0$ on the RoboTwin 2.0 benchmark \cite{chen2025robotwin}.

\subsection{Evaluation on False-Completion Benchmark}
\label{sec: eval-fc}

\noindent \textbf{Quantitative Results.}
Table \ref{table:false} presents a comparative evaluation of existing VLA methods on False-Completion benchmark.
Both ReViP and ReViP* achieve the highest average success rate and the best average rank across all eight tasks, outperforming the strongest baseline, $\pi_0$-Fast, by 18\% and the baseline model, $\pi_0$, by 26\%. 
This aggregate improvement demonstrates that rebalancing visual and proprioceptive inputs with visual cues consistently enhances performance across diverse perturbations.

In Object Drop tasks, where policies must detect failures rather than follow an inertial plan, ReViP achieves 62.4\% across five drop tasks, with ReViP* reaching 65.2\%.
This demonstrates that TSO enhances visual failure detection and replanning, preventing the blind continuation of proprioception-driven progress. 
In Distractor Swap tasks, where object recognition and manipulation are confused, ReViP substantially improves over $\pi_0$ (15\% $\rightarrow$ 37\%), with ReViP* reaching 39\%, highlighting progress-aware visual cues'role in disambiguating similar instances. 
In Relayout task, ReViP achieves 84\% with ReViP* improving to 88\%, surpassing $\pi_0$ (70\%) and UniVLA (20\%).
These results underscore ReViP*'s superior performance and validate the benefit of higher VLM capacities in enhancing cue quality, aligning with our objective of rebalancing vision and proprioception.

\begin{table}[t]
    \caption{\textbf{Experimental Results of False-Completion Benchmark.} Success rates (SR) and average ranks (RK) across eight tasks grouped into three perturbation sources: Object Drop, Distractor Swap, and Relayout.
    $^{\clubsuit}$ indicates non-tabletop cabinet scenes (i.e., cabinet scenes). The best and second-best results are shown in \textbf{bold} and \underline{underline}, respectively.
    Our method achieves the highest performance across the benchmark.
    }
    \label{table:false}
    \centering
    \scriptsize
    \setlength{\tabcolsep}{2.5pt}
    \renewcommand{\arraystretch}{0.95}
    \resizebox{\columnwidth}{!}{%
    \begin{tabular}{l *{5}{c} | *{2}{c} | c | cc}
        \toprule
        \multirow{4}{*}{\textbf{Methods}} & \multicolumn{5}{c}{\textbf{Object-Drop}} & \multicolumn{2}{c}{\textbf{Distractor-Swap}} & \textbf{Relayout} & \multicolumn{2}{c}{\textbf{Average}}\\
       & Butter  & Cheese & Bottle & Carton & $\text{Bowl}^{\clubsuit}$ & Butter & Cheese & Bowl-Plane & \multirow{4}{*}{ SR $\uparrow$} & \multirow{4}{*}{RK $\downarrow$}  \\ 
       &  \raisebox{-0.01\height}{\includegraphics[width=0.042\linewidth]{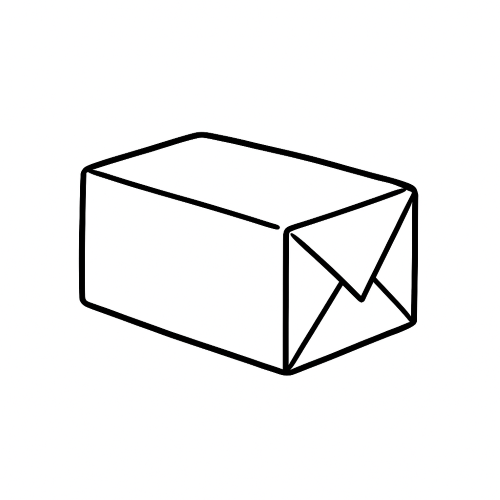}}
       &  \raisebox{-0.01\height}{\includegraphics[width=0.048\linewidth]{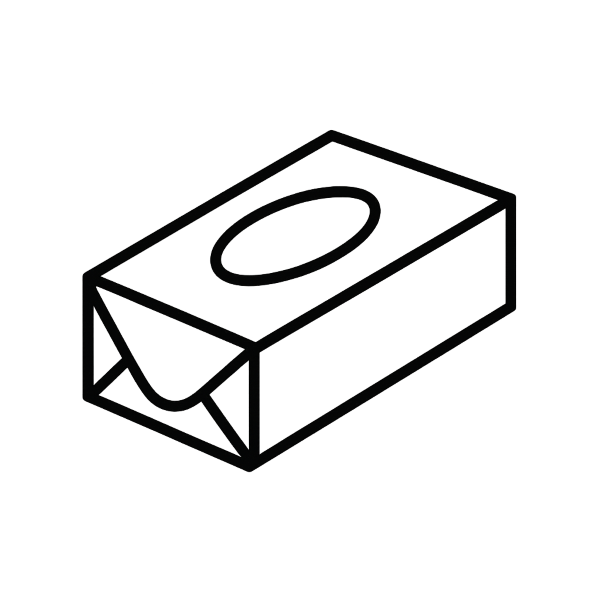}}
       &  \raisebox{-0.01\height}{\includegraphics[width=0.054\linewidth]{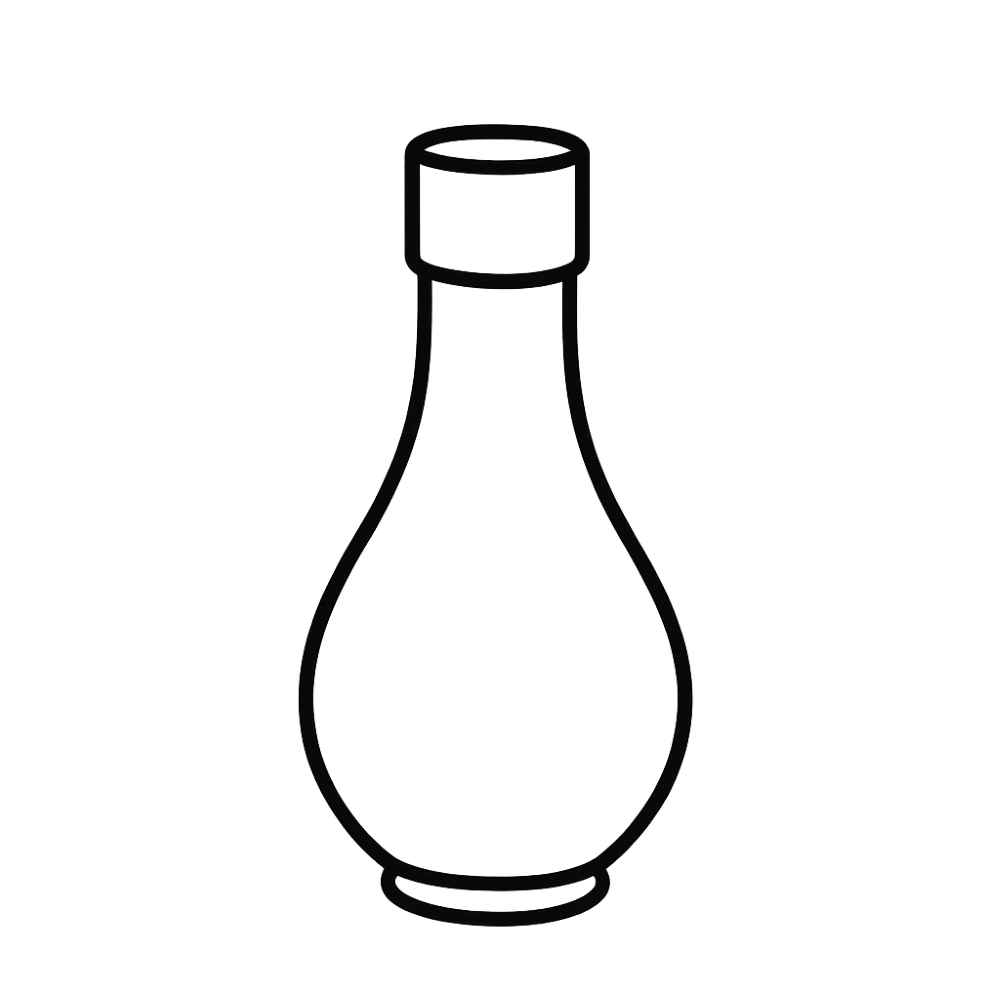}}
       &  \raisebox{-0.01\height}{\includegraphics[width=0.054\linewidth]{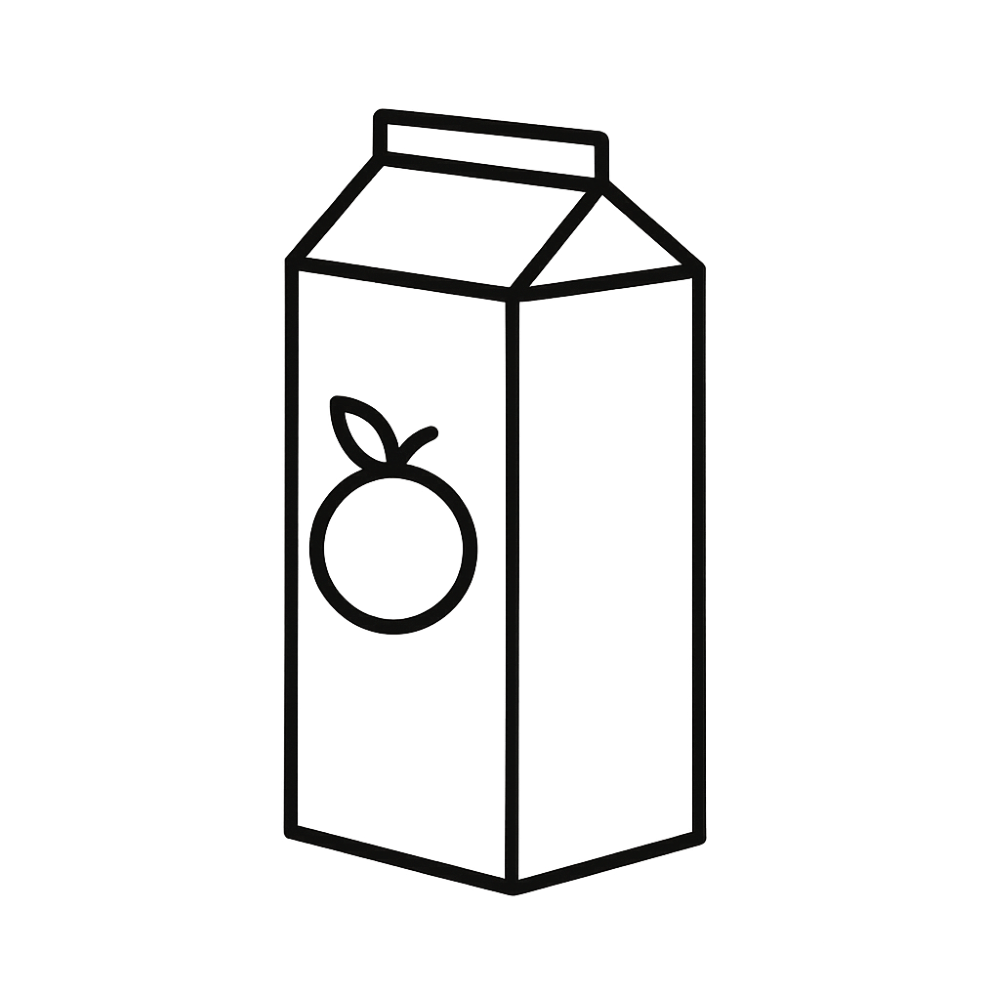}}
       &  \raisebox{-0.01\height}{\includegraphics[width=0.054\linewidth]{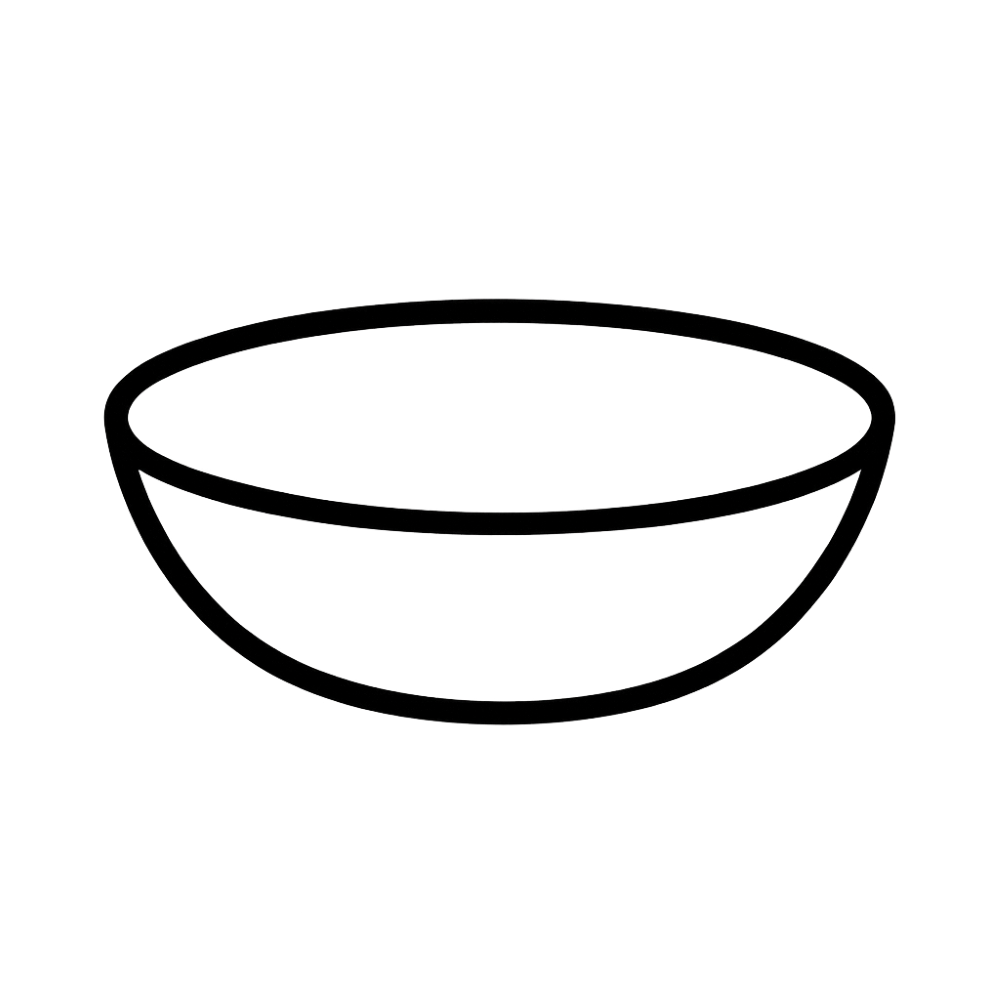}}
       &  \raisebox{-0.01\height}{\includegraphics[width=0.042\linewidth]{Figures/FC/Butter.png}}
       &  \raisebox{-0.01\height}{\includegraphics[width=0.048\linewidth]{Figures/FC/Cheese.png}}
       &  \raisebox{-0.08\height}{\includegraphics[width=0.060\linewidth]{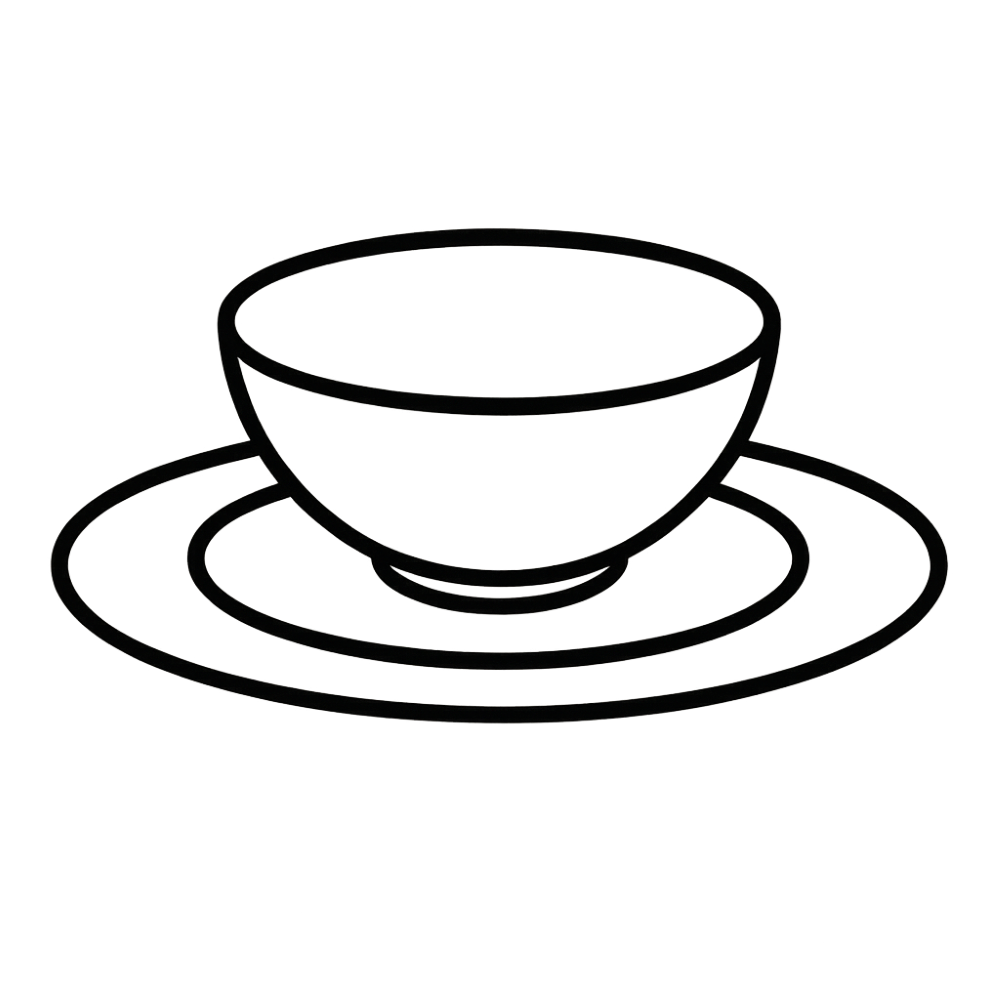}} 
       &  \\
        \midrule
        OpenVLA ~\cite{kim2024openvla} & 12\% & 30\% & 16\% & \phantom{0}6\% & \phantom{0}2\% & \phantom{0}0\% & \phantom{0}0\% & 12\% & 10\% & 7\\
        $\boldsymbol{\pi_0}$~\cite{black2024pi_0}   & 24\% & 78\% & 26\% & 40\% & 20\% & \phantom{0}6\% &24\% & 70\% & 36\% & 4\\
        $\boldsymbol{\pi_0}$-Fast~\cite{fast} & 28\% & 62\% & 20\% & 16\% & \textbf{58\%} & \textbf{36\% } & 44\% & \underline{84\%} & 44\% & 3\\  
        OpenVLA-OFT~\cite{oft} & 16\% & 42\% & 18\% & 12\% & 35\% & 2\% & 10\% & 18\% & 19\% & 5\\
        UniVLA~\cite{univla}  & 18\% & 40\% & 16\% & \phantom{0}10\% & \phantom{0}6\% & \phantom{0}4\% & 12\% & 20\% & 18\% & 6\\
        \rowcolor{gray!20} ReViP & \underline{46\%} & \underline{92\%} & \underline{66\%} & \underline{62\%} & \underline{46\%} & 26\% &\textbf{48\%} &\underline{84\%} & \underline{59\%} & 2\\
        \rowcolor{gray!20} ReViP*  & \textbf{50\%} & \textbf{96\%} & \textbf{72\%} & \textbf{66\%} & 42\% & \underline{32\%} &\underline{46\%} &\textbf{88\%} & \textbf{62\%} & 1\\
        \bottomrule
    \end{tabular}%
    }
    \vspace{-6mm}
\end{table}

\begin{figure*}[t]
    \centering
    \includegraphics[width=0.9\linewidth]{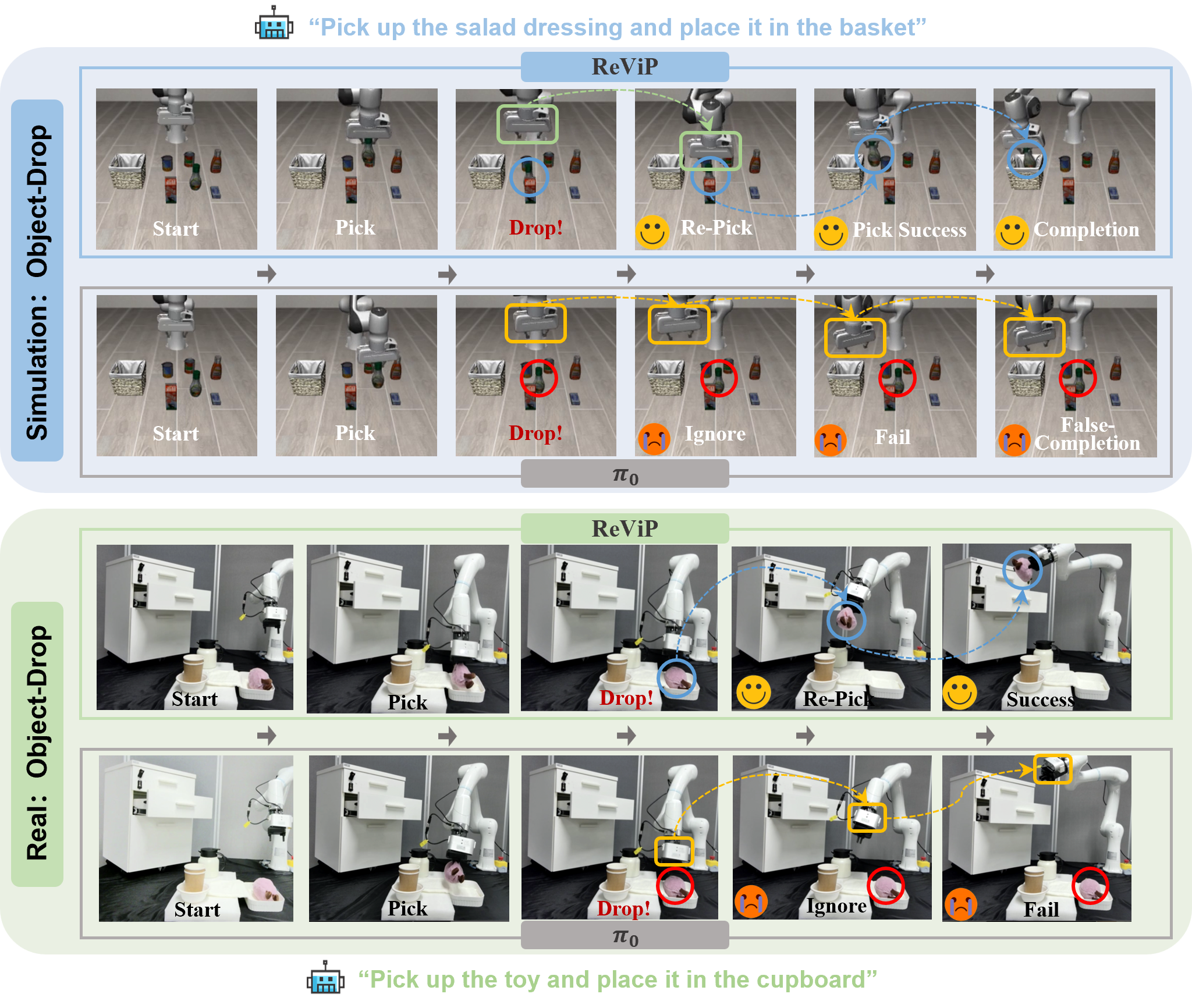}
    \caption{Qualitative comparisons on our False Completion benchmark with Object Drop settings (above: simulation, below: real-world). For simulation, ReViP detects the object drop during execution, and successfully re-picks the target (salad dressing), achieving a true completion $\smiley$, while $\pi_0$ fails to react to the clear visual failure and continues executing a state-dominant bias, resulting in false completion $\frownie$. In the real world, ReViP can also retrieve the toy when it falls, but $\pi_0$ still fails to respond correctly.}
    \label{fig:FC_results}
    \vspace{-5mm}
\end{figure*}  
\subsection{Evaluation on General Simulation Benchmark}
\label{sec: eval-general}

\noindent \textbf{Qualitative Results.}
Fig.~\ref{fig:FC_results} presents a representative comparison between ReViP and $\pi_0$ under the Object Drop perturbation. 
When the bottle slips from the gripper, $\pi_0$ continues its proprioception-driven trajectory, ignoring the visible failure and ultimately declaring a false completion. 
In contrast, ReViP immediately responds to the drop event, triggers a re-pick attempt, and guides the policy to replan, recover the object, and successfully complete the task. 
This qualitative behavior highlights the mechanism behind ReViP's performance gains on False-Completion Benchmark: decisions are based on the current visual observation rather than on internal proprioceptive progress.

Along with the numerical improvements in Table \ref{table:false}, these qualitative findings show that ReViP consistently transforms failure cases into successful completions through vision-proprioception rebalancing.

\noindent \textbf{Performance on LIBERO.} As shown in Table \ref{table:libero}, we evaluate ReViP on the LIBERO benchmark, including LIBERO-Spatial, LIBERO-Object, LIBERO-Goal, and LIBERO-10 (also called LIBERO-Long), each with 10 tasks and 50 demonstrations.
Both ReViP and ReViP* consistently outperform all other models, achieving the highest average task success rate of 96.7\%, surpassing UniVLA and the backbone model $\pi_0$.
Specifically, on LIBERO-Spatial, ReViP reaches 99.0\% success, effectively saturating the suite.
On the more challenging LIBERO-10 suite which involves long-horizon manipulation across mixed task families, ReViP improves $\pi_0$'s success rate from 85.2\% to 89.8\%, and ReViP* further reaches 92.2\%, demonstrating better stability over extended executions.

\begin{wraptable}[9]{r}{0.52\columnwidth}
\vspace{-11mm}
\centering
\caption{\textbf{Challenging Dual-Arm RoboTwin 2.0 (Easy$\rightarrow$Hard) Results.} Success rates (\%), best in \textbf{bold}.}
\label{tab:robotwin}
\scriptsize
\setlength{\tabcolsep}{3pt}
\resizebox{0.95\linewidth}{!}{%
\begin{tabular}{lccc >{\columncolor{gray!15}}c}
\toprule
\textbf{Task} & \textbf{DP3} & \textbf{RDT} & {\boldmath$\pi_0$} & \textbf{ReViP} \\
\midrule
Place Object Stand  & 0  & 5  & 11 & \textbf{22} \\
Place Bread Basket  & 1  & 2  & 4  & \textbf{15} \\
Pick Dual Bottles   & 1  & 13 & 12 & \textbf{21} \\
Put Bottles Dustbin & 21 & 4  & 13 & \textbf{24} \\
\midrule
\textbf{Average}    & 6  & 6  & 10  & \textbf{21} \\
\bottomrule
\end{tabular}%
}
\vspace{-2mm}
\end{wraptable}
These results indicate that the proposed vision and proprioception rebalance with visual cues not only mitigates false completion but also improves performance on standard manipulation tasks.
\begin{table*}[t]
    \caption{\textbf{Experimental Results of Simulation Experiments on the LIBERO benchmark.} Success rates (SR) and average ranks (RK) across four task suites: Spatial, Object, Goal, and 10.
    The best and second-best results are shown in \textbf{bold} and \underline{underline}, respectively.
    Our method achieves the highest performance across all suites.}
    \label{table:libero}
    \vspace{-2mm}
    \centering
    \resizebox{0.8\columnwidth}{!}{%
    \begin{tabular}{l *{5}{|cc}}
        \toprule
        \multirow{2}{*}{\textbf{Methods}} & \multicolumn{2}{c}{\textbf{Spatial}} & \multicolumn{2}{c}{\textbf{Object}} & \multicolumn{2}{c}{\textbf{Goal}} & \multicolumn{2}{c}{\textbf{10}} & \multicolumn{2}{c}{\textbf{Average}}\\
        & SR $\uparrow$ & RK $\downarrow$ & SR $\uparrow$ & RK $\downarrow$ & SR $\uparrow$ & RK $\downarrow$ & SR $\uparrow$ & RK $\downarrow$ & SR $\uparrow$ & RK $\downarrow$ \\
        \midrule
        SpatialVLA \cite{spatialvla}   & 88.2 &9 & 89.9 &9 & 78.6&10 & 55.5&11 & 78.1 &10\\
        $\boldsymbol{\pi_0}$ \cite{black2024pi_0}     & 96.8 &3 & \textbf{98.8} &1  & 95.8&3 & 85.2&6 & 94.2 &5 \\
        $\boldsymbol{\pi_0}$-Fast \cite{fast}     & 96.4 & 5 & 96.8 &5  & 88.6&7 & 60.2&10 & 85.5&8 \\
        CoT-VLA \cite{cotvla}                   & 87.5 & 10 & 91.6 &8  & 87.6&8 & 69.0&9 & 83.9 &9\\
        STAR \cite{star}                     & 95.5 & 6 & 98.3 &3& 95.0 &5& 88.5&4 & 94.3 & 4\\
        $\boldsymbol{\pi_{0.5}}$-KI \cite{pi05}  & \underline{98.0} & 2 & 97.8 &4 & 95.6&4  & 85.8&5 & 94.3 &4 \\
        TriVLA    \cite{trivla}            & 91.2 & 8 & 93.8&6  & 89.8&6  & 73.2&8 & 87.0 &6 \\
        GR00T N1.5 \cite{gr00t}          & 92.0 & 7 & 92.0 & 7 & 86.0 & 9 & 76.0 &7 & 86.5 & 7 \\
        UniVLA \cite{univla}                    & 96.5 & 4 & 96.8 &5 & 95.6 &4& \underline{92.0}&2 & 95.2 &3 \\
        \rowcolor{gray!20} ReViP         & \textbf{99.0} & 1 & \underline{98.6} &2 & \underline{96.2} &2 & 89.8 & 3 & \underline{95.9} &2\\
        \rowcolor{gray!20} ReViP*         & \textbf{99.0} & 1 & \textbf{98.8} &1 & \textbf{96.6} &1 & \textbf{92.2} & 1 & \textbf{96.7} &1\\
        \bottomrule
    \end{tabular}}
\end{table*}

\noindent\textbf{Performance on RoboTwin 2.0 (Dual Arm).} 
To further evaluate ReViP in dual-arm scenarios, we test on the RoboTwin 2.0 benchmark, featuring contact-rich manipulation (Table~\ref{tab:robotwin}).
We consider four representative tasks, each trained with 50 demonstrations collected in clean environments.
Evaluation under \textit{hard mode conditions} includes 100 trials per task with domain randomization, lighting variations, object clutter, and table height shifts.
ReViP achieves the highest success rates, outperforming RDT and $\pi_0$, indicating the task-stage feedback mechanism scales effectively to dual-arm tasks and remains robust under severe environmental perturbations.

\subsection{Real-World Experiment}
\label{sec: real-world}

\begin{wrapfigure}{r}{0.5\linewidth}
    \vspace{-8mm}
    \centering
    \includegraphics[width=0.8\linewidth]{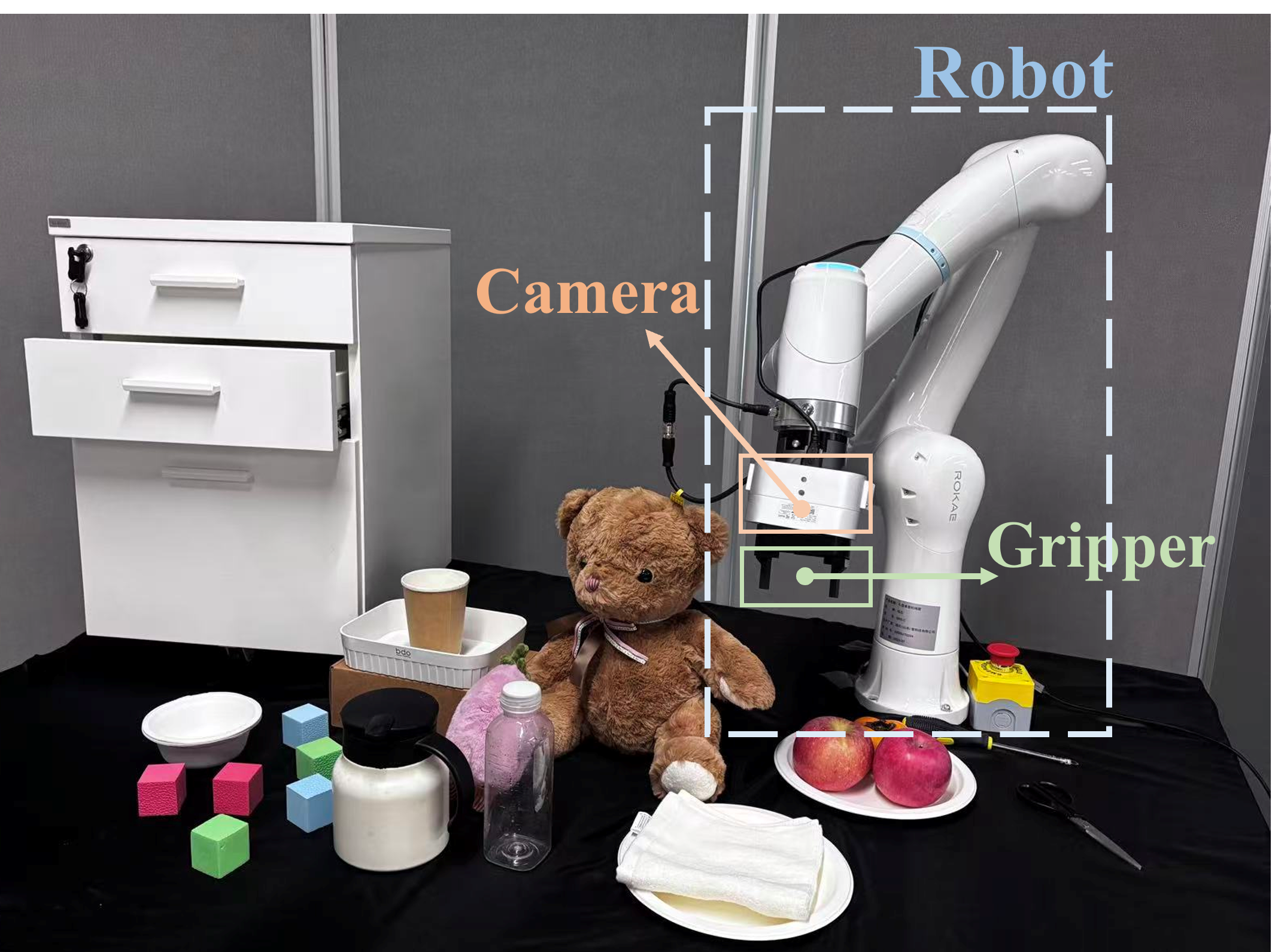}
    \caption{Real-world experiments use a robotic arm with grippers and cameras.} 
    \label{fig:real_setting}
    \vspace{-6mm} 
\end{wrapfigure}

\noindent \textbf{Robot platform.} As shown in Fig.~\ref{fig:real_setting}, our real-world experiments use a 6-dof ROKAE robotic arm equipped with a JODELL gripper. 
Two Orbbec Femto Bolt RGB-D cameras capture both first-person and third-person views.

During data collection, the robot is guided through kinesthetic teaching, with trajectories manually generated and recorded via interpolated playback.

\noindent\textbf{Tasks and Datasets.} 
Our datasets include two manipulation tasks with three different perturbations: Drop, Swap, and Relayout.
(1) \textit{Pick-and-Place}: This task evaluates VLA robustness against visual distractions, with multiple distractor objects requiring accurate target localization and manipulation.
(2) \textit{Drawer-Manipulation}: This task tests the VLA's ability to handle long-horizon sequences, involving dynamic scene interactions and a series of actions in the correct order. 
Further details are in the supplementary materials.


\noindent\textbf{Baseline.} 
We compare ReViP with two SOTA methods, $\boldsymbol{\pi_0}$ and $\boldsymbol{\pi_0}$-Fast, to assess its performance in handling multi-distractor scenarios and its error-correction ability in long-horizon tasks.

\begin{table*}[t]
    \caption{\textbf{Experimental Results of Real-World Experiments.} Success rates (SR) are reported. The best result is shown in \textbf{bold}.} 
    \label{tab:real-world}
    \vspace{-2mm}
    \centering
    \resizebox{0.68\columnwidth}{!}{%
    \begin{tabular}{l *{2}{c} | *{4}{c} }
        \toprule
        \multirow{2}{*}{\textbf{Methods}} & \multicolumn{2}{c}{\textbf{Object Placement}} & \multicolumn{3}{c}{\textbf{Drawer Manipulation}} &\textbf{Average}  \\
        & Cup $\rightarrow$ Box & +Drop & Toy $\rightarrow$ Drawer & +Close & +Drop  & SR $\uparrow$ \\
        \midrule
        $\boldsymbol{\pi_0 }$ \cite{black2024pi_0}   & 8/10 & 4/10 & 7/10 & 9/10  & 3/10 & 62\%  \\
        $\boldsymbol{\pi_{0}}$-Fast \cite{fast}     & 7/10 & 3/10 & 7/10 & 8/10  & 3/10& 56\% \\
        \rowcolor{gray!20} ReViP  & \textbf{10/10} & \textbf{8/10}  & \textbf{9/10} & \textbf{10/10} & \textbf{7/10} & \textbf{88\%} \\
        \bottomrule
    \end{tabular}
    }
    \vspace{-6mm}
\end{table*}

\noindent\textbf{Results.} 
As shown in Table \ref{tab:real-world} and Fig.~\ref{fig:FC_results}, ReViP achieves the highest success rate in both sub-tasks and overall performance. 
Under disturbances, ReViP detects the dropped object and autonomously re-grasps it, showing successful completion.

\subsection{Extended Real-World Evaluation}
We further evaluate ReViP under new, challenging real-world conditions to test robustness against false completion. The extended tasks include high-contrast backgrounds, alternative cameras (e.g., RealSense), small objects, and long-horizon multi-step sequences. 
These extended results (Table~\ref{tab:supp_real-world} and Fig.~\ref{fig:setting}) confirm ReViP's robustness under challenging settings. Detailed task designs, scene and hardware variations, and qualitative results are in the supplementary materials.

\subsection{Ablation Study}
\label{sec:Ablation Study}
\noindent\textbf{Effect of VLMs for TSO.} 
\begin{table*}[!t]
    \caption{
    \textbf{Extended Real-World Evaluation: New, Challenging Settings}
    Success rates (SR) are reported. The best result is shown in \textbf{bold}.
    }
    \label{tab:supp_real-world}
    \vspace{-2mm}
    \centering
    \resizebox{0.8\columnwidth}{!}{%
    \begin{tabular}{l *{3}{c} | *{2}{c} | {c} | {c} }
        \toprule
        \multirow{2}{*}{\textbf{Methods}} & \multicolumn{3}{c}{\textbf{Cup Placement}} & 
        \multicolumn{2}{c}{\textbf{Cube Placement}} &
        {\textbf{Drawer}} &\textbf{Average}  \\
        & Cup $\rightarrow$ Box & +Drop & +Distractor 
        & Cube $\rightarrow$ Box & +Distractor 
        & \textbf{Manipulation} & SR $\uparrow$ \\
        \midrule
        $\boldsymbol{\pi_0 }$ \cite{black2024pi_0}   
        & 5/10 & 4/10 & 3/10 
        & 4/10 & 2/10 
        & 4/10 & 34\%  \\
        $\boldsymbol{\pi_{0}}$-Fast \cite{fast} 
        & 3/10 & 1/10 & 2/10 
        & 2/10 & 0/10 
        & 6/10 & 23\%  \\
        \rowcolor{gray!20} ReViP  
        & \textbf{9/10} & \textbf{8/10}  & \textbf{6/10} 
        & \textbf{7/10} & \textbf{6/10} 
        & \textbf{8/10} & \textbf{73}\% \\
        \bottomrule
    \end{tabular}}
    \vspace{-2mm}
\end{table*}

To assess TSO's robustness, we evaluate various VLM backbones (Table \ref{table:vlms}), including LLaVA-1.5-7B \cite{liu2023llava}, InternVL2.5-8B~\cite{chen2024internvl}, and Qwen2.5-VL~\cite{Qwen2.5-VL}.
Performance is consistent across VLMs, indicating TSO is not tied to a specific backbone. Stronger VLMs generally lead to higher success rates, with Qwen2.5-VL-72B achieving the best average SR (62\%, +26\%).

\begin{table}[ht]
    \caption{\textbf{Ablation Study: Effect of VLMs for TSO.} 
    Success rates (SR) on the False-Completion benchmark are reported. $^{\clubsuit}$ indicates non-tabletop cabinet scenes.}
    \label{table:vlms}
    \vspace{-2mm}
    \centering
    \scriptsize
    \setlength{\tabcolsep}{2.5pt}
    \renewcommand{\arraystretch}{0.95}
    \resizebox{\columnwidth}{!}{%
    \begin{tabular}{l *{5}{c} | *{2}{c} | c | c}
        \toprule
        \multirow{2}{*}{\textbf{Methods}} & \multicolumn{5}{c}{\textbf{Object-Drop}} & \multicolumn{2}{c}{\textbf{Distractor-Swap}} & \textbf{Relayout} & {\textbf{Average}}\\
       & Butter  & Cheese & Bottle & Carton & $\text{Bowl}^{\clubsuit}$ & Butter & Cheese & Bowl-Plane & SR $\uparrow$ \\ 
        \midrule
        Baseline~\cite{black2024pi_0}   & 24\% & 78\% & 26\% & 40\% & 20\% & \phantom{0}6\% & 24\% & 70\% & 36\%\\
        LLaVA-1.5-7B \cite{liu2023llava}  & 32\% & 94\% & 70\% & 36\% & 64\% & 14\% & 40\% & 74\% &  55\% (\textcolor{green!50!black}{$\uparrow$ 19\%})\\
        InternVL2.5-8B~\cite{chen2024internvl} & {48\%} & 80\% & {72\%} & {64\%} & {60\%} & 16\% & 32\% & 80\% & 57\% (\textcolor{green!50!black}{$\uparrow$ 21\%})\\  
        Qwen2.5-VL-3B~\cite{Qwen2.5-VL} & 46\% & {92\%} & {66\%} & 62\% & {46\%} &{26\%} & f{48\%} & {84\%} & {59\%} (\textcolor{green!50!black}{$\uparrow$ 23\%})\\
        Qwen2.5-VL-72B~\cite{Qwen2.5-VL}  & {50\%} & {96\%} & {72\%} & {66\%} & 42\% & {32\%} & {46\%} & {88\%} & {62\%} (\textcolor{green!50!black}{$\uparrow$ 26\%})\\
        \bottomrule
    \end{tabular}%
    }
    \vspace{-4mm}
\end{table}

\begin{wraptable}{r}{0.52\columnwidth}
\vspace{-5mm}
\centering
\caption{\textbf{Ablation Study: FiLM factor $\alpha$.} Success rates (SR) on the False-Completion-Drop benchmark are reported.}
\label{tab:factor}
\scriptsize
\setlength{\tabcolsep}{3pt}
\resizebox{0.92\linewidth}{!}{%
\begin{tabular}{lccccc}
    \toprule
    $\alpha$ & 0 & 0.25 & 0.5 & 1.0 & Learnable \\
    \midrule
    SR $\uparrow$ & 38.6\% & 56.4\% & 62.4\% & 46.8\% & 65.2\% \\
    \bottomrule
\end{tabular}%
}
\vspace{-8mm}
\end{wraptable}

\noindent\textbf{Effect of FiLM factor $\alpha$.} 
Table~\ref{tab:factor} shows that fixed $\alpha$ performs reasonably across a range of values, while learnable $\alpha$ in ReViP adaptively selects modulation strength and achieves higher performance, 
supporting the effectiveness of our designs.

\subsection{Comparison of Efficiency and Performance}
\label{sec:Method Efficiency}
\begin{wraptable}[7]{r}{0.52\columnwidth}
\vspace{-11mm}
\centering
\caption{\textbf{Comparison of efficiency and performance.} 
Real-time latency, control frequency, and success rate (SR) on the False-Completion benchmark are reported.}
\label{tab:efficiency}
\scriptsize
\setlength{\tabcolsep}{3pt}
\resizebox{\linewidth}{!}{%
\begin{tabular}{lccc}
    \toprule
    \textbf{Methods} & \textbf{Latency (ms)} $\downarrow$ & \textbf{Frequency (Hz)} $\uparrow$ & \textbf{SR (\%)} $\uparrow$ \\
    \midrule
    OpenVLA-OFT~\cite{oft} & 175 & \phantom{0}5.71 & 19 \\
    $\boldsymbol{\pi_0}$~\cite{black2024pi_0} & {44.6} & {22.42} & 36 \\
    \rowcolor{gray!20} ReViP & 62.4 & 16.03 & {59} \\
    \bottomrule
\end{tabular}%
}
\end{wraptable}

Table \ref{tab:efficiency} reports real-time efficiency and performance on our False-Completion benchmark using a single NVIDIA H100 GPU (30 warm-up runs and mean of the next 50).
OpenVLA-OFT is slow and inaccurate, while $\pi_0$ is fast but lower in performance. 
ReViP achieves a favorable efficiency--performance balance with modest overhead.
TSO is triggered once per action chunk and pipelined asynchronously, ensuring a control frequency of 16 Hz, sufficient for real-time operation with only slight latency compared to $\pi_0$.



\label{sec: Generality}

\section{Conclusion}
In this paper, we identify \emph{false completion} as a critical failure mode in VLA models that conflicts with human commonsense task evaluation, and systematically study modality imbalance to reveal its causes. 
We further introduce the first False-Completion benchmark suite for evaluating false completion, and propose \emph{ReViP}, a VLA framework that adaptively rebalances vision and proprioception using visual cues. 
Extensive simulation and real-world experiments demonstrate that ReViP effectively mitigates false completion, highlighting the benefits of vision-proprioception rebalancing.



%
%
\bibliographystyle{splncs04}
\bibliography{main}

\appendix   
\clearpage
\setcounter{page}{1}

   {
   \newpage
        \centering
        \Large
        \vspace{0.5em}Supplementary Material \\
        \vspace{1.0em}
   }

This appendix provides comprehensive supplementary material to support the methodology, analysis, and findings presented in the main paper. It includes implementation details covering model architectures and training configurations, along with additional experimental results and analyses.


\section{Additional Implementation Details}
\subsection{Model Details}

\noindent \textbf{Task-Stage Observer.}
The Task-Stage Observer (TSO) is built upon the Qwen 2.5 VL model~\cite{Qwen2.5-VL} and is designed to perform task-relevant reasoning
that first identifies the robot’s currently visible physical state and the
spatial location and status of the task-relevant objects, then summarizes the immediate stage intention aligned with the given goal.
The TSO takes three inputs: the task instruction, the task goal prompt, and a third-person observation image that covers the full workspace.
This global viewpoint enables reliable retrieval of object-level evidence even under perturbations.

To ensure that the TSO produces concise and structured semantics rather than free-form scene descriptions, we design a dedicated instruction protocol for the model, as shown in Table~\ref{tab:instruction}.
This protocol guides the model to enumerate visual evidence, specify stage-wise completion conditions, and infer an explicit task stage description that is aligned with the ongoing trajectory. Following the hidden state aggregation process detailed in Section~3.2 of the main paper, the TSO output is transformed into a single embedding vector of dimension $2048$ for ReViP ($8192$ for ReViP*). This vector serves as the task stage representation $z_t$ and is subsequently fed into the Task-Stage Enhancer to enable task stage feature-wise modulation in ReViP.

\begin{table*}[t]
\centering
\caption{Instruction details for the Task-Stage Observer.}
\label{tab:instruction}
\renewcommand{\arraystretch}{1.3}
\begin{tabular}{p{0.22\linewidth} p{0.75\linewidth}}
\toprule
\textbf{Component} & \textbf{Content} \\
\midrule

\textbf{System Prompt} &
You are an expert robot task assistant, integrated with a robot arm capable of executing physical actions in the real world.
Based on the task goal and observation, provide the additional task-stage information to the robot in a structured and precise way.\\

\textbf{Instruction Rules} &
\begin{itemize}
    \item LIST the current state of all relevant objects (robot arm, scene objects) as visual evidence.
    \item DEFINE clear completion conditions for each task stage.
    \item Determine the current \texttt{task\_stage\_cues} based on object states and describe it as an ongoing action.
\end{itemize} \\

\textbf{Input Fields} &
Task Goal: \verb|{task_goal}| \\
&
Observation: \verb|{Third person view image}| \\

\textbf{Output Format} & 
\verb|{ "visual_evidence": ["<evidence_1>", "..."],| \\
\verb|                   "task_stage_cues": "<task_stage_cues>"} |\\ 
\bottomrule
\end{tabular}
\end{table*}

\noindent \textbf{Task-Stage Enhancer.}
In ReViP, the Task-Stage Observer outputs a task stage representation $z_t \in \mathbb{R}^{d}$, which is processed by a lightweight bottleneck network to produce the modulation parameters $(\gamma_t, \beta_t)$.
The mapping $h(\cdot)$ is implemented as a two-layer MLP with SiLU activations.
To maintain stable training, the final projection uses a Xavier uniform initializer scaled by $0.01$ with zero-initialized biases, so that TS-FiLM behaves close to identity at the early stage while the learnable scalar $\alpha$ gradually adjusts the modulation strength.
The resulting $\gamma_t, \beta_t \in \mathbb{R}^{D}$ are broadcast across tokens and applied to the concatenated vision and language prefix $P_t$ following Eq.~(5) in the main paper.
Importantly, the binary prefix mask $M_t \in \{0,1\}^{B \times S}$ is applied after modulation to ensure that TS-FiLM only affects valid prefix tokens and fully preserves the padding structure used by the attention backend.

\noindent \textbf{Attention Mechanism.}
After applying TS-FiLM to obtain the modulated prefix $\tilde P_t$, the tokens are fed into the vision language backbone together with a suffix sequence that contains the state token and the diffusion-conditioned noisy action tokens.
Let $S_t \in \mathbb{R}^{B \times 1 \times D}$ denote the state token and $A_t \in \mathbb{R}^{B \times H \times D}$ denote the sequence of action and timestep tokens.
Then, the action generator's input $X_t$ can be defined as follows:
\begin{equation}
X_t \;=\; [\,\tilde P_t;\ S_t;\ A_t\,] \in \mathbb{R}^{B \times L \times D},
\end{equation}
where $L = S + 1 + H$ is the total token length.

Self attention is then applied over $X_t$ with a block structured attention mask $\mathcal{M}_t^A$ that encodes the prefix style interaction pattern inherited from the VLA backbone:
\begin{equation}
H_t \;=\; \mathrm{Attn}\bigl(Q(X_t),\,K(X_t),\,V(X_t);\ \mathcal{M}_t^A\bigr).
\end{equation}
Prefix tokens in $\tilde P_t$ attend to each other across vision and language, but do not attend to the suffix tokens.
In contrast, the suffix tokens in $S_t$ and $A_t$ are allowed to attend to all prefix tokens as well as to the appropriate subset of previous suffix tokens according to the autoregressive part of $\mathcal{M}_t^A$.
The binary prefix mask $M_t$ is applied after TS-FiLM so that only valid prefix positions contribute to $Q$, $K$, and $V$, and padding tokens remain completely excluded.
This design lets the diffusion head integrate task stage modulated visual context with proprioceptive state and time-dependent action information, while keeping the original routing and padding structure of the attention mechanism unchanged during both training and inference.

\subsection{Training Details.}
\noindent \textbf{Training Setup.}
We adopt $\pi_0$~\cite{black2024pi_0} as the backbone model with full parameter fine-tuning and set the action chunk size to $K = 50$.
The model is trained for 60k steps with a batch size of 32.
The learning rate follows a cosine decay schedule with 3{,}000 warmup steps, increasing to a peak of $2.0\times10^{-5}$ and then decaying to a final learning rate of $2.0\times10^{-6}$.
We use the AdamW optimizer with gradient clipping set to 1.0 and apply an exponential moving average (EMA) with a decay value of 0.999.

To avoid excessive training latency caused by online task-stage extraction, we preprocess the entire dataset using the Task-Stage Observer before training. 
For every action step, we extract the corresponding task-stage intention and store it as part of the training annotations. 
During training, the progress-aware visual cues are injected into ReViP to provide explicit task-stage guidance, enabling the model to more effectively rebalance visual and proprioceptive features. 
At inference time, task-stage intention is computed online by the Task-Stage Observer and fed into ReViP in real time.

\section{Additional Experiments Details}
\subsection{False Completion Benchmark Suite Details}
\begin{table*}[ht]\small
\centering
\caption{
Overview of the False Completion Benchmark Suite.
The suite contains eight LIBERO manipulation tasks constructed from three perturbation families: Object Drop, Distractor Swap, and Object Relayout.
For each perturbation type, we provide representative visual examples together with the corresponding language instructions.
The target object is highlighted in \textcolor{purple!90}{purple} and the goal region is highlighted in \textcolor{teal!90}{teal}.
}
\label{tab:false_completion_tasks}
\resizebox{0.95\linewidth}{!}{%
\begin{tabular}{m{0.19\linewidth} | m{0.6\linewidth} | m{0.22\linewidth}}
\toprule 
Type & Instruction & Visual Example \\
\midrule

\multirow{9}{*}{Object Drop}
& \multirow{9}{*}{%
    \parbox{\linewidth}{%
        \ding{182} Pick up the \textcolor{purple!90}{butter} and place it in the \textcolor{teal!90}{basket} \\
        \ding{183} Pick up the \textcolor{purple!90}{cream cheese} and place it in the \textcolor{teal!90}{basket} \\
        \ding{184} Pick up the \textcolor{purple!90}{salad dressing} and place it in the \textcolor{teal!90}{basket} \\
        \ding{185} Pick up the \textcolor{purple!90}{orange juice} and place it in the \textcolor{teal!90}{basket} \\
        \ding{186} Pick up the \textcolor{purple!90}{black bowl} in the \textcolor{teal!90}{top drawer} of the wooden cabinet and place it on the plate
    }%
}
& \multirow{9}{*}{%
    \centering
    \includegraphics[height=0.21\textwidth, keepaspectratio]{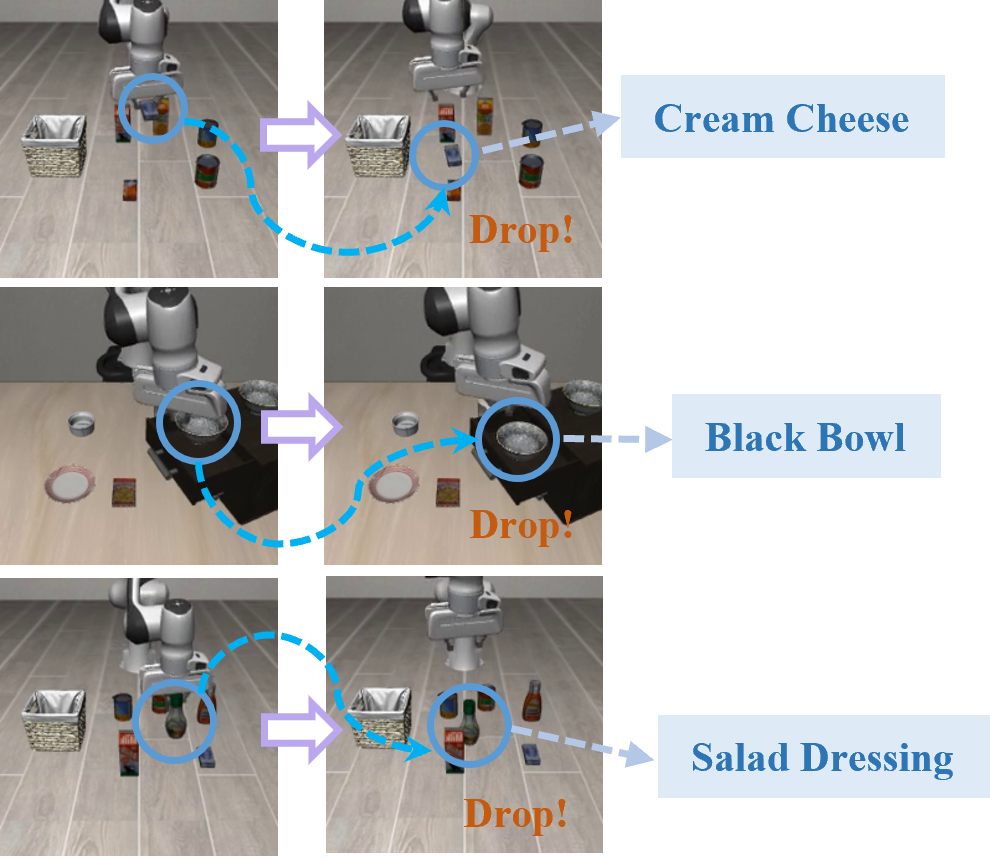}%
}
\\
& & \\  
& & \\  
& & \\  
& & \\  
& & \\  
& & \\  
& & \\  
& & \\  




\midrule

\multirow{3}{*}{Distractor Swap}
& \multirow{3}{*}{%
    \parbox{\linewidth}{%
        \ding{187} Pick up the \textcolor{purple!90}{butter} and place it in the \textcolor{teal!90}{basket}.\\
        \ding{188} Pick up the \textcolor{purple!90}{cream cheese} and place it in the \textcolor{teal!90}{basket}.%
    }%
}
& \multirow{3}{*}{%
    \centering
    \includegraphics[height=0.07\textwidth, keepaspectratio]{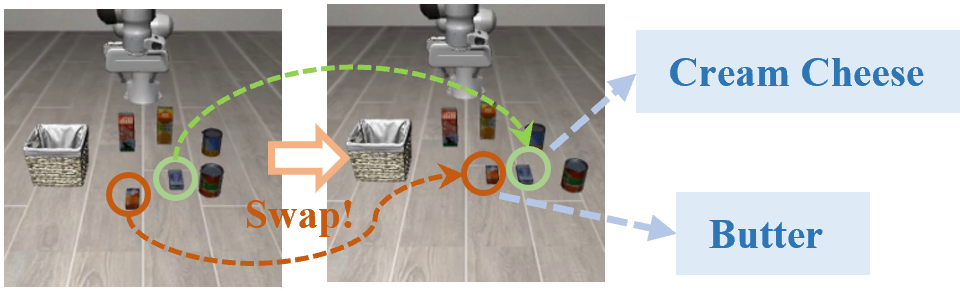}%
}
\\
& & \\  
& & \\  

\midrule

\multirow{3}{*}{Object Relayout}
& \multirow{3}{*}{%
    \parbox{\linewidth}{%
        \ding{189} Pick up the \textcolor{purple!90}{black bowl} next to the plate and place it on the \textcolor{teal!90}{plate}.
    }%
}
& \multirow{3}{*}{%
    \centering
    \includegraphics[height=0.07\textwidth, keepaspectratio]{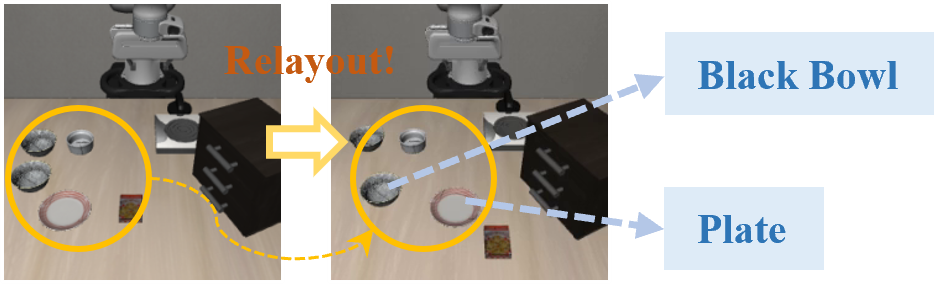}%
}
\\
& & \\  
& & \\  
\bottomrule
\end{tabular}
}
\end{table*}
False completion arises when external disturbances or unintended robot actions invalidate the ongoing plan while the robot still declares success.
To study state-dominant bias and visually apparent false completions in a controlled manner, we construct a False Completion Benchmark Suite on top of LIBERO that injects targeted perturbations and probes complementary aspects of the phenomenon.

We consider three perturbation sources that commonly induce false completion: \textit{Object Drop}, \textit{Distractor Swap}, and \textit{Object Relayout}.
Together, these perturbations form a suite of \textbf{8 manipulation tasks} that constitute the proposed False Completion Benchmark.
The instruction remains fixed, and only the environment state is perturbed either at the start of the episode or during execution.
This disturbance design prevents success via replaying demonstrations like proprioceptive trajectories and instead compels policies to rely on current visual evidence for progress checking and replanning.
An overview of all tasks, including representative visualizations and the corresponding language instructions, is provided in Table~\ref{tab:false_completion_tasks}.
Below, we detail the three perturbation families.

\noindent \textbf{Object Drop.}
This setting tests recovery from unexpected displacement and exposes false completion most directly.
We instantiate three variants that differ in object scale and contact context, which in total yield five tasks:
\begin{itemize}
    \item \emph{Small Object Drop:} a piece of cheese and a block of butter are grasped and then forcibly dropped onto the table, requiring the policy to visually re-detect the fallen small object and regrasp it.
    \item \emph{Large Object Drop:} a bottle of salad dressing and a carton of orange juice are lifted and then dropped, which introduces longer trajectories and larger motion amplitude compared with small object cases.
    \item \emph{Bowl Drop in Cabinet:} a bowl is dropped inside a cabinet with occlusion and contact constraints, and the policy needs to visually localize the fallen bowl and complete the task under clutter and limited visibility.
\end{itemize}

\noindent \textbf{Distractor Swap.}
This setting evaluates whether the policy detects that two visually similar objects have exchanged positions rather than replaying a proprioception-driven trajectory.
The initial poses of the target and a visually similar distractor are swapped while the instruction remains fixed.
Concretely, cheese and butter exchange locations after an initial movement phase, and the policy must complete the original instruction with respect to the correct target object.
This produces two tasks that differ only in which object is specified as the linguistic target.

\noindent \textbf{Object Relayout.}
This setting breaks demonstration-specific spatial priors while preserving the linguistic target.
The target object and its goal region are jointly relocated to new feasible positions within the workspace.
The policy is therefore required to update its plan based on the current visual configuration rather than relying on stale internal state or memorized spatial layouts.
In our benchmark, this gives one relayout task that relocates both the target and its goal surface.

Overall, the three perturbation families jointly emphasize visual grounding, continuous progress monitoring, and effective replanning under distribution shifts that frequently result in false completions.
The suite provides a focused and diagnostic evaluation framework for VLA models, isolating failure modes that arise when policies overly trust internal state progression in the presence of visually apparent disturbances.

\subsection{Task Definitions and Evaluation Settings}

For all tasks in the False Completion Benchmark Suite, an episode is considered successful only when the agent completes the instructed goal within the predefined step budget. Any premature termination or incorrect placement is counted as failure, even if the policy internally declares success.

\noindent \textbf{Step Budget.}
The three perturbation families differ in whether the disturbance introduces additional action requirements.
For Distractor Swap and Object Relayout, the perturbation only alters the initial scene configuration without increasing manipulation complexity.
The number of required actions remains comparable to the original LIBERO tasks, and therefore, the maximum episode length is kept unchanged.
In contrast, Object Drop requires the agent to detect the unexpected displacement, relocate the fallen object, and perform an additional re-pick action.
This introduces extra motion beyond the demonstration regime.
To ensure that policies have sufficient budget to perform recovery, we extend the maximum step limit by an additional 30 to 50 steps, depending on the specific drop variant.

\noindent \textbf{Drop Trigger Rule.}
For Object Drop tasks, we design a deterministic and reproducible drop triggering mechanism to guarantee fair comparisons across models.
A drop event is initiated only when the agent maintains a continuous gripper closing state for a fixed duration, indicating that the object has been securely grasped and lifted.
Once this condition is satisfied, the environment forces the gripper to open, causing the object to fall from a consistent spatial location.
This rule ensures that all models experience the same drop timing and position, removes stochasticity, and isolates the evaluation of false completion behavior to perception and decision making rather than random disturbances.

\noindent \textbf{Fairness Across Perturbations.}
For Distractor Swap and Object Relayout, perturbations are applied deterministically at environment initialization.
Since the instruction remains unchanged and the agent starts from the same initial state distribution, all models operate under identical scene configurations.
For Object Drop, the deterministic drop trigger further ensures that each model receives an identical disturbance during execution.
Together, these settings guarantee that the benchmark provides a controlled and strictly fair evaluation of policies under visually apparent distribution shifts that may induce false completion.

\subsection{Supplementary Simulation Study: State Masking on LIBERO}
To complement the real-robot modality-mismatch study in the main paper, we conduct a modality-ablation experiment in simulation on LIBERO~\cite{libero} to further validate our findings. 
LIBERO covers 40 diverse manipulation tasks across four sub-suites, providing a representative large-scale testbed for evaluating the effect of modality ablation under a unified protocol (Section \ref{sec:libero}). 
This supplementary experiment extends the empirical support for our conclusion beyond the real-world setting and tests whether the same trend holds across diverse manipulation scenarios.

\begin{wraptable}{r}{0.5\columnwidth}
\vspace{-4mm}
\centering
\caption{\textbf{Experimental Results of existing VLA models under modality control on LIBERO.} Success Rates (\%) are reported.}
\label{tab:libero_modality}
\scriptsize
\setlength{\tabcolsep}{3pt}
\resizebox{\linewidth}{!}{%
\begin{tabular}{lccccc}
    \toprule
    \textbf{Methods} & \textbf{Spatial} $\uparrow$ & \textbf{Object} $\uparrow$ & \textbf{Goal} $\uparrow$ & \textbf{10} $\uparrow$ & \textbf{Average} $\uparrow$\\
    \midrule
    $\boldsymbol{\pi_0}$ \cite{black2024pi_0}     & 96.8 & 98.8 & 95.8 & 85.2 & 94.2 \\
    w/ State-Masked & 94.8 (\textcolor{red}{$\downarrow$ 2.0}) & 95.8 (\textcolor{red}{$\downarrow$ 3.0}) & 90.6 (\textcolor{red}{$\downarrow$ 5.2}) & 80.6 (\textcolor{red}{$\downarrow$ 4.6}) & 90.5 (\textcolor{red}{$\downarrow$ 3.7})\\
    \bottomrule
\end{tabular}%
}
\vspace{-5mm}
\end{wraptable}

As shown in Table~\ref{tab:libero_modality}, the simulation results exhibit a trend consistent with the real-world observations.
When the state input is directly masked, performance drops across all four sub-suites. Specifically, compared with the original $\pi_0$, the average success rate decreases from 94.2\% to 90.5\%, with declines of 2.0\% on Spatial, 3.0\% on Object, 5.2\% on Goal, and 4.6\% on LIBERO-10.
These results suggest that although removing state input can reduce the policy’s reliance on proprioceptive cues, it also removes task-relevant information that is important for normal task execution. Notably, the degradation is more pronounced on Goal and LIBERO-10, indicating that proprioceptive information remains particularly important for tasks requiring stable progress tracking and accurate action execution.

Overall, the simulation study is consistent with the real-robot analysis: naively removing state inputs may partially alleviate state-dominant behavior, but it introduces a clear cost to task performance. 
These findings further strengthen \textbf{our motivation for ReViP}, suggesting that \emph{addressing false completion should shift from directly removing state inputs to rebalancing visual and proprioceptive signals, so as to preserve the complementary benefits of multimodal information for robust task execution.}


\subsection{LIBERO \& RoboTwin 2.0 Benchmark Details}
\label{sec:libero}
\noindent \textbf{LIBERO Simulation Benchmark.}
We evaluate ReViP in the LIBERO simulation benchmark~\cite{libero}, a standardized collection of language-conditioned robotic manipulation tasks designed to probe multimodal grounding, visual reasoning, and action generation. Compared with earlier platforms such as RLBench \cite{rlbench}, LIBERO provides richer language instructions, more diverse scene layouts, and a broader range of object-level interactions, making it a more suitable testbed for evaluating vision language action models.

LIBERO consists of four curated task suites: Spatial, Object, Goal, and Long. Each suite contains ten manipulation tasks accompanied by 50 human teleoperated demonstrations. These suites are designed to decouple different aspects of visuomotor reasoning:
\begin{itemize}
\item \textbf{LIBERO Spatial} tests spatial reasoning by placing identical objects in varying spatial configurations. Success requires interpreting relations such as left, right, in front of, or behind as described in the instruction.
\item \textbf{LIBERO Object} evaluates generalization across object categories. While spatial layouts are fixed, the target and distractor objects vary in type, shape, and color, requiring accurate object-level grounding from both vision and language.
\item \textbf{LIBERO Goal} probes goal-oriented understanding by varying the goal specification while holding object identities and spatial layouts constant. Fine-grained semantic distinctions in instructions must be mapped to distinct manipulation outcomes.
\item \textbf{LIBERO Long} introduces long-horizon tasks involving multiple intermediate steps, diverse objects, and broader spatial compositions. Successful execution requires perception grounding, semantic tracking, and sequential planning.
\end{itemize}

ReViP is trained and evaluated under the same standard LIBERO protocol as OpenVLA \cite{kim2024openvla} and $\pi_0$ to ensure comparability. By covering spatial reasoning, object identification, goal disambiguation, and long-horizon sequencing, the LIBERO benchmark offers a comprehensive and controlled environment for assessing the robustness and visual grounding capabilities of vision-language action models. This makes LIBERO an appropriate platform for evaluating the contributions of task stage modulation in ReViP.

\noindent \textbf{RoboTwin 2.0 Benchmark.}
RoboTwin 2.0~\cite{chen2025robotwin} is a scalable simulation framework and benchmark for robust bimanual robotic manipulation.
It instantiates 50 dual-arm tasks across five robot embodiments and provides a large-scale dataset of clean expert trajectories.
At the core of RoboTwin 2.0 is a diverse object library and an automatic expert data synthesis pipeline that uses multimodal language models and simulation in the loop refinement to generate realistic task-level execution code.
To enhance sim-to-real transfer and robustness, RoboTwin 2.0 applies structured domain randomization along five axes, including clutter, lighting, background, tabletop height, and language instructions, and defines unified evaluation protocols for dual arm policies\cite{chen2025robotwin}.

In our study, RoboTwin 2.0 serves as a complementary benchmark to LIBERO and is used to assess whether ReViP extends effectively from single-arm settings to more challenging dual-arm coordination.
We select four representative tasks from the 50-task suite that cover typical bimanual interaction patterns and object configurations.
Following the official RoboTwin 2.0 protocol, we fine-tune each model with 50 clean expert demonstrations per task and evaluate it with 100 rollouts under the Hard setting with domain randomization.
This evaluation regime imposes a strong robustness requirement and provides a stringent test of the compared vision language action policies.

\subsection{Real-World Setup}
Our real-world experiments are performed on a 6-DoF ROKAE robotic arm equipped with a JODELL parallel-jaw gripper.
We collect real-world successful demonstration trajectories for the Robot Arm via manually moving the robot along the desired path. 
For all tasks, we gather 50 expert demonstrations. 
At each control step, the policy receives a main image and wrist image and a 7-dimensional proprioceptive state in joint space (6 joint positions and 1 gripper scalar). The model outputs action joints in an action chunk of 50.

The main paper reports results on two representative manipulation Tasks 1-2, while this appendix extends the evaluation with additional extended Tasks 3-5 designed to assess hardware robustness, scene generalization, and multi-step reasoning ability (please see Section \ref{sec: additional_real_world}). 

The instructions and descriptions for Tasks 1-2 are provided below: \begin{itemize}
\item  ``\textit{Pick up the cup on the white plate, then place it inside the white plastic square box}''

This is a structured pick-and-place scenario where the plate serves as a visually distinctive anchor for the target object.
The task evaluates basic visual grounding and manipulation stability. In the main paper, this task further includes Drop perturbations to assess the model’s ability to recover from false-completion cases.

\item  ``\textit{Pick up the toy and place it in the open drawer, then close the drawer}''

A multi-stage drawer manipulation task requiring correct sequencing: grasping, placing, and completing the closing motion.
This task tests long-horizon reasoning, stage transitions, and re-localization when object visibility changes near drawer boundaries.

\end{itemize}

\noindent \textbf{Drop Trigger Mechanism}
We evaluate the model’s response to false-completion scenarios using a controlled Drop perturbation.
To ensure \textbf{repeatability, fairness, and naturalistic failure behavior}, we implement Drop using a hardware-in-the-loop condition:

First, during execution, we continuously monitor the gripper width in real time.
Once the gripper closes beyond a threshold, indicating that it has successfully grasped the object, the robot proceeds with the placement trajectory.
Then, at a predefined early stage of this motion, we command the gripper to open, causing the object to fall naturally along its original motion path.

\section{Supplementary Quantitative Analysis}

\subsection{Extended Real-World Task Results}
\label{sec: additional_real_world}
Beyond the standard configuration described above, the appendix includes additional hardware and scene generalization experiments to further evaluate the robustness of ReViP under real-world variations not covered in the main paper. To this end, we introduce three additional tasks that involve new backgrounds, different camera hardware, new object categories, and extended multi-step action sequences. In particular, we replace the wrist-mounted cameras with the ORBBEC DaBai DCW module and substitute the tabletop background with a high-contrast green cloth. These changes alter imaging characteristics, depth quality, and background complexity, creating a significantly more challenging perception environment.

The instructions and descriptions for the extended Tasks 3-5 are provided below:\begin{itemize}
\item \textit{“Pick up the cup on the table, then place it inside the white plastic square box”}.
This task removes the plate anchor used in the main paper and employs a plain tabletop decorated with a high-contrast green cloth. The modified background introduces strong color bias and lower object-background separability, making target localization more difficult.

\item \textit{“Pick up the small red cube on the table, then place it inside the white box”}.
A fine-grained manipulation task involving a small, low-profile cube whose visual features are harder to extract under depth noise. Surrounding distractors further increase the challenge, requiring precise visual grounding and accurate grasping.

\item \textit{“Pick up the purple toy on the table, then place it into the middle drawer of the cabinet, and finally close the drawer”}.
This extended long-horizon task involves grasping, precise placement, and drawer closing. The scene contains up to ten objects of diverse shapes and colors, creating substantial visual clutter. This setting rigorously tests the model’s ability to maintain stable grounding and execute multi-step actions under complex external interference.
\end{itemize}

Table~6 in the main paper summarizes the success rates across all extended tasks. The results reveal several consistent trends:

\noindent \textbf{Cup Placement.}
ReViP substantially outperforms both $\pi_0$ and $\pi_0$-Fast across all variants, especially under Drop and Distractor settings. The improvement highlights ReViP’s stronger ability to recover from false-completion risks (e.g., object drop) and to reject visually similar distractors in cluttered scenes.

\noindent \textbf{Cup Placement.}
The cube task poses significant difficulty due to the small size and weak geometric features of the target object. While both baselines suffer heavily under distractor interference, ReViP maintains considerably higher success rates, demonstrating improved fine-grained localization and stable visual grounding.

\noindent \textbf{Drawer Manipulation.}
ReViP also achieves the highest performance on the long-horizon drawer task. Baseline policies frequently exhibit state-dominant behavior, prematurely closing the drawer after empty grasps or proceeding despite misalignment. In contrast, ReViP more reliably detects errors, re-localizes the target, and completes the full multi-step sequence.

Across all real-world settings, ReViP maintains substantially higher robustness under perturbations and complex visual clutter. These results confirm that the proposed vision-proprioception rebalance improves both false-completion resilience and long-horizon execution in challenging, unconstrained environments.


\subsection{Study of the Model’s Plug-and-Play Capability}
\begin{table}[t]
    \caption{\textbf{Experimental Results on the Plug-and-Play Capability of Vision-Proprioception Rebalance.} Success rates (SR) across two task suites:
     Object Drop in False Completion benchmarks \& LIBERO 10.
    ReViP and ReViP$_{0.5}$ consistently improve success rates across both suites.
    $^{*}$ indicates results from from the 
    official GitHub implementation.
    }
    \label{tab:revip}
    \centering
    \begin{tabular}{l *{2}{c}}
        \toprule
        \multirow{1}{*}{\textbf{Methods}} & \textbf{Object-Drop} & \textbf{LIBERO-10} \\
        \midrule
        $\boldsymbol{\pi_0}$ \cite{black2024pi_0}  & 37.6\% & 85.2\% \\
        $\boldsymbol{\pi_{0.5}}^{*}$ \cite{pi05}    & 54.4\% & 92.4\% \\
        \rowcolor{gray!20} ReViP    & 65.2\% (\textcolor{purple}{$\uparrow$ 27.6\%}) & 92.2\% (\textcolor{purple}{$\uparrow$ 7.0\%})\\
        \rowcolor{gray!20} ReViP$_{0.5}$ & 68.2\% (\textcolor{purple}{$\uparrow$ 13.8\%}) & 95.8\% (\textcolor{purple}{$\uparrow$ 3.4\%})\\
        \bottomrule
    \end{tabular}
\end{table}


To further evaluate the plug-and-play capability of the \textit{Vision-Proprioception Rebalance} beyond the $\pi_0$ \cite{black2024pi_0} backbone used in ReViP, we extend our framework to another state-of-the-art VLA architecture, $\pi_{0.5}$ \cite{pi05}. We instantiate ReViP on $\pi_{0.5}$, denoted ReViP$_{0.5}$ by integrating the same Task-Stage Observer, Task-Stage Enhancer, and TS-FiLM pipeline, without adding any extra modules or backbone-specific tuning. This setup examines whether the proposed progress-aware visual cues injection functions in a plug-and-play manner across different policy formulations and whether the benefits observed in ReViP naturally transfer to new architectures.

For evaluation, we adopt two challenging task suites that probe complementary aspects of robustness:
\textit{(i) Object Drop}, containing five representative false-completion tasks requiring failure recovery, and
\textit{(ii) LIBERO-10}, a long-horizon manipulation suite demanding consistent grounding and precise action sequencing.
Both suites remain identical across backbones to ensure fair comparison.

As shown in Table~\ref{tab:revip}, ReViP and ReViP${0.5}$ consistently improve success rates across both task suites on their respective backbones, $\pi{0}$ and $\pi_{0.5}$. On the $\pi_0$ backbone, ReViP achieves improvements of 27.6\% on Object Drop and 7.0\% on LIBERO-10. When transferred to $\pi_{0.5}$, the same design further yields gains of 13.8\% on Object Drop and 7.0\% on LIBERO-10. These results demonstrate that task-stage conditioning through structured visual evidence functions effectively in a plug-and-play fashion across different VLA architectures, indicating that the proposed vision-proprioception rebalance mechanism is broadly applicable and \textbf{not tied to any specific backbone.}

\section{Supplementary Qualitative Analysis}
We present additional qualitative results from both simulation and real-world experiments, focusing on challenging false-completion scenarios and demonstrating ReViP’s improved robustness, generalization, and execution quality.

\subsection{Additional Visualizations of Simulation Results}

\begin{figure*}[t]
    \centering
    \includegraphics[width=0.85\linewidth]{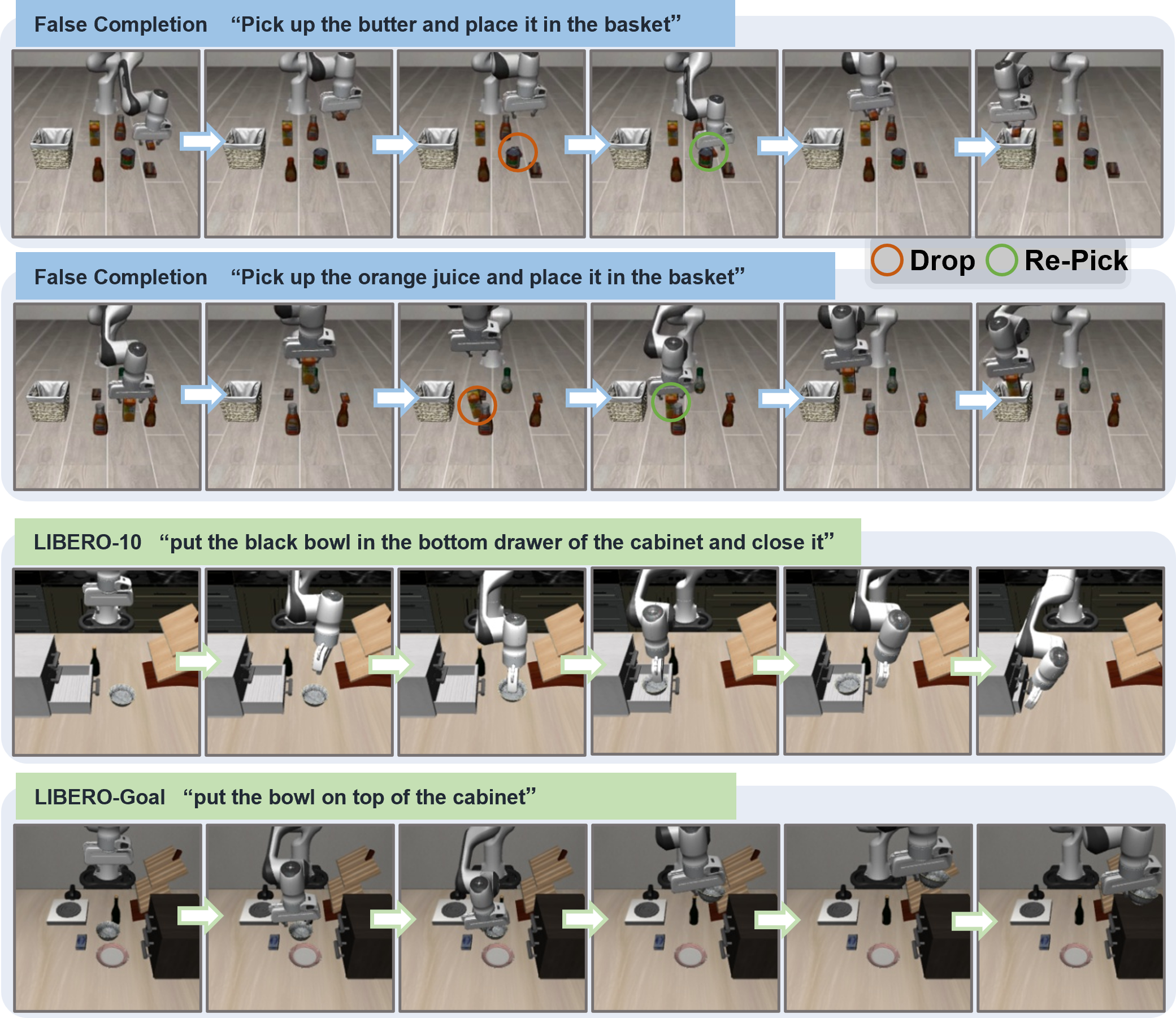}
    \caption{
    Additional qualitative results in simulation on the False-Completion Benchmark, LIBERO-10, and LIBERO-Goal. 
    In the False-Completion Benchmark, ReViP detects the object drop during execution and re-grasps the target, achieving true completion instead of prematurely terminating.
    On LIBERO-10 and LIBERO-Goal, ReViP consistently completes long-horizon tasks under diverse disturbances, environments, and object configurations.
    These visualizations highlight the complexity of the tasks and further demonstrate ReViP’s robustness, generalization, and execution capabilities.
    }
    \label{fig:simulations}
    \vspace{-4mm}
\end{figure*}

\begin{figure*}[!h]
    \centering
    \includegraphics[width=0.92\linewidth]{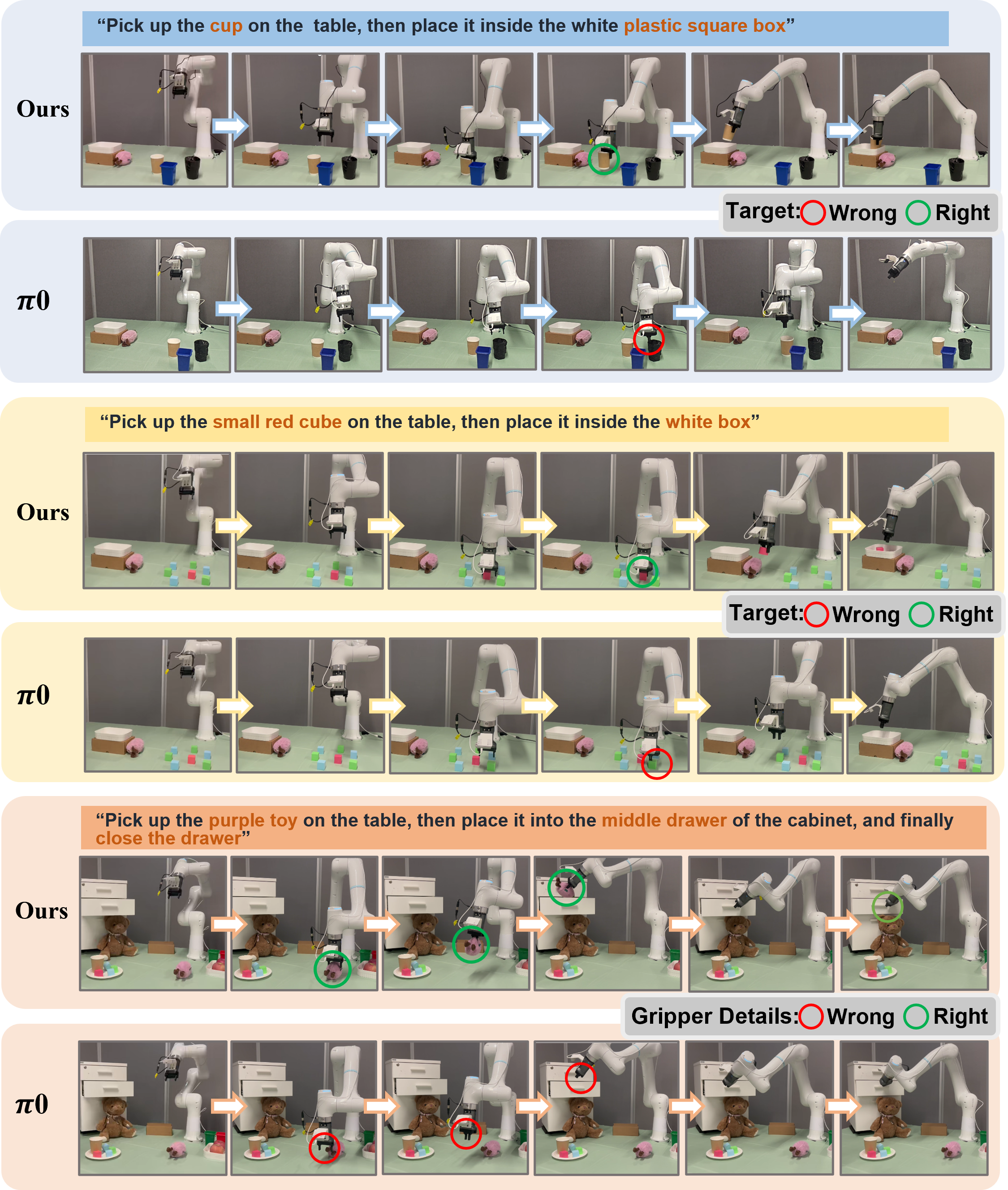}
    \caption{
    Additional qualitative real-world comparisons between ReViP and $\pi_{0}$ on challenging false-completion scenarios and long-horizon manipulation tasks. 
    ReViP accurately identifies the correct target in cluttered scenes and avoids mis-grasp-induced false completion, while $\pi_{0}$ is easily confused by distractors. 
    For long-horizon drawer tasks, ReViP re-localizes the target and completes the full sequence, whereas $\pi_{0}$ often proceeds with incorrect state-driven actions. 
    The corresponding videos are provided in the supplementary materials.
    }
    \label{fig:supp_real_world}
    \vspace{-4mm}
\end{figure*}

As shown in Figure \ref{fig:simulations}, we present additional simulation results on both the false-completion benchmark and LIBERO. The model consistently completes multi-step tasks under diverse false-completion disturbances, environments, and object configurations. These visualizations highlight the complexity of the tasks and further demonstrate ReViP’s generalization and execution capabilities.

\subsection{Additional Visualizations of Real-World Results}
Figure \ref{fig:supp_real_world} provides qualitative comparisons between ReViP and the baseline $\pi_{0}$ across two representative real-world settings: false-completion scenarios and long-horizon manipulation tasks. The visualizations highlight how ReViP maintains correct visual grounding and robust task progression.

In cluttered environments containing many distractors with highly similar appearance, ReViP accurately identifies the true target object, such as distinguishing the intended cup or red cube from visually similar items. This allows ReViP to execute the correct grasping action. In contrast, $\pi_{0}$ is frequently misled and grasps an incorrect object, which triggers an incorrect stage transition and results in a typical false-completion failure despite clear visual evidence of mis-grasp.

For complex long-horizon manipulation tasks, $\pi_{0}$ often follows internal state progression rather than visual feedback. This leads to errors such as pushing or closing the cabinet even when the gripper is empty, ultimately causing task failure. ReViP detects these inconsistencies, re-localizes the missing target, and then completes the full sequence of grasping, placing, and cabinet operations.

These qualitative examples show that ReViP mitigates false-completion behaviors and significantly improves long-horizon task reliability through better integration of visual observations. The corresponding real-world videos are provided in the supplementary materials.


\end{document}